\documentclass{article} 
\usepackage{iclr2026_conference,times}
\iclrfinalcopy

\usepackage{amsmath,amsfonts,bm}

\def\eqref#1{equation~\ref{#1}}

\def\1{\bm{1}}

\DeclareMathAlphabet{\mathsfit}{\encodingdefault}{\sfdefault}{m}{sl}
\SetMathAlphabet{\mathsfit}{bold}{\encodingdefault}{\sfdefault}{bx}{n}

\usepackage[linesnumbered,ruled,vlined]{algorithm2e}
\usepackage[utf8]{inputenc} 
\usepackage[T1]{fontenc}    
\usepackage{hyperref}       
\usepackage{url}            
\usepackage{booktabs}       
\usepackage{amsfonts}       
\usepackage{nicefrac}       
\usepackage{microtype}      
\usepackage{xcolor}         

\usepackage{booktabs}       
\usepackage{colortbl}

\usepackage{graphicx}
\usepackage{float}                
\usepackage{subfig}             
\usepackage{overpic}
\usepackage{enumitem}
\usepackage{multirow}
\usepackage{amsthm}
\usepackage{wrapfig}
\usepackage{array}
\usepackage{picinpar}
\usepackage{wrapfig}

\title{FracAug: Fractional Augmentation boost Graph-level Anomaly Detection under Limited Supervision}

\author{Xiangyu Dong \\
        The Chinese University of Hong Kong \\
        \texttt{xydong@se.cuhk.edu.hk} \\
    \And
        Xingyi Zhang \\
        Mohamed bin Zayed University of Artificial Intelligence \\
        \texttt{xingyi.zhang@mbzuai.ac.ae} \\
    \And
        Sibo Wang\thanks{corresponding author}\\
        The Chinese University of Hong Kong \\
        \texttt{swang@se.cuhk.edu.hk} \\
}
\newcommand{\vect}[1]{\boldsymbol{#1}}

\newtheorem{theorem} {Theorem}

\newtheorem{proposition} {Proposition}

\begin{document}

\maketitle

\begin{abstract}
Graph-level anomaly detection (GAD) is critical in diverse domains such as drug discovery, yet high labeling costs and dataset imbalance hamper the performance of Graph Neural Networks (GNNs). To address these issues, we propose FracAug, an innovative plug-in augmentation framework that enhances GNNs by generating semantically consistent graph variants and pseudo-labeling with mutual verification. Unlike previous heuristic methods, FracAug learns semantics within given graphs and synthesizes fractional variants, guided by a novel weighted distance-aware margin loss. This captures multi-scale topology to generate diverse, semantic-preserving graphs unaffected by data imbalance. Then, FracAug utilizes predictions from both original and augmented graphs to pseudo-label unlabeled data, iteratively expanding the training set. As a model-agnostic module compatible with various GNNs, FracAug demonstrates remarkable universality and efficacy: experiments across 14 GNNs on 12 real-world datasets show consistent gains, boosting average AUROC, AUPRC, and F1-score by up to 5.72\%, 7.23\%, and 4.18\%, respectively.

\end{abstract}

\section{Introduction}
\label{sec:introduction}
Graph-structured data is pivotal in real applications ranging from drug discovery to anomaly identification among proteins \citep{igad22zhang}. While Graph Neural Networks (GNNs) excel at modeling topological and feature-based patterns through message-passing, their effectiveness in Graph-level Anomaly Detection (GAD)---distinguishing anomalous graphs from normal ones---is hindered by two key challenges: limited supervision and extreme class imbalance, as demonstrated in Section \ref{sec:experiments}. Specifically, anomalies represent rare instances, exacerbating data imbalance and restricting the availability of labeled training samples \citep{consisgad24chen, rqgnn24dong}. While data augmentation techniques have revolutionized computer vision \citep{imageaug23zhang} by generating synthetic labels through rotations or crops, their adaptation to graph domains presents unique challenges. Unlike images, graphs inhabit non-Euclidean space where seemingly minor structural modifications (e.g., edge removal) risk distorting semantic properties and violating the label-invariant assumption---a critical constraint in GAD’s challenging setting of limited supervision and inherent class imbalance. 

Existing graph-level augmentation methods, such as MAA \citep{maa22yoo}, often employ heuristic modifications without considering data properties, leading to compromised semantics or insufficient diversity in GAD tasks. Consequently, their direct application may underperform vanilla GAD models. We attribute this gap to three key issues: (1) the absence of semantic-preserving augmentation strategies, (2) inadequate handling of imbalance, and (3) ineffective utilization of unlabeled data.

To address these challenges, we introduce FracAug, a novel plug-in augmentation framework that generates semantic-preserving graph variants and pseudo-labels for unlabeled graphs to train GNNs for GAD. Our key innovation leverages the fractional power of adjacency matrices, which encodes multi-scale topological relationships. By computing polynomials of various fractional graphs, guided by weighted distance-aware margin loss, FracAug introduces controlled structural variations while ensuring semantic consistency with the original graph’s label, independent of the underlying data distribution. Afterward, a given GNN will produce predictions for both original and synthetic samples, enabling FracAug to employ a mutual verification mechanism for pseudo-labeling unlabeled graphs, thereby iteratively expanding the training set. This approach not only mitigates supervision scarcity but also enhances model robustness against class imbalance.

In summary, our contributions are as follows:

\begin{itemize}[topsep=0.5mm, partopsep=0pt, itemsep=0pt, leftmargin=10pt]
    \item We present FracAug, the first augmentation framework designed for GAD that maintains effectiveness under the dual constraints of limited supervision and imbalanced distribution
    \item FracAug operates as a model-agnostic plug-in augmentation framework compatible with 14 GNNs without architectural modifications, facilitating seamless integration into existing models. 
    \item Extensive experiments on 12 real-world datasets demonstrate that FracAug enhances performance across diverse GNNs, significantly outperforming existing graph augmentation approaches.
 \end{itemize}

\section{Related Work}
\label{sec:relatedwork}

{\bf Graph Classification.} Generalized GNNs, such as GCN \citep{gcn17kipf}, GraphSAGE \citep{graphsage17hamilton}, GAT \citep{gat18velickovic}, and GIN \citep{gin19xu}, excel at learning graph representations through neighborhood aggregation. Recent advances include LRGNN \citep{lrgnn23wei}, which captures long-range dependencies with stacking GNNs, and GRDL \citep{grdl24wang}, which achieves state-of-the-art (SOTA) performance by learning representation distributions of graphs. However, these representative GNNs are not specifically designed for GAD tasks. While they can capture certain topological or feature-based patterns, their performance degrades under data imbalance and limited supervision. 

{\bf Graph-level Anomaly Detection.} Recognizing the challenges underlying GAD tasks, researchers have introduced specialized approaches to address them. For instance, iGAD \citep{igad22zhang} introduces dual-discriminative kernels guided by a point mutual information-based loss function to better capture graph anomalies. Later, by mapping anomalies and normal graphs to separate areas based on adjusted candidate nodes, GmapAD \citep{gmapad23ma} shows advanced performance. Moreover, RQGNN \citep{rqgnn24dong} leverages the Rayleigh Quotient to detect graph anomalies effectively within spectral space. Recently, UniGAD \citep{unigad24lin} combines different levels of graph anomaly detection to capture comprehensive information to enhance the detection accuracy. Although these specialized frameworks show promising performance in addressing data imbalance challenges, they still present inferior results due to the limited supervision issue. 

{\bf Graph-level Augmentation.} 
To address the scarcity of labeled examples, researchers also develop diverse augmentation techniques for graph-level tasks. For example, MAA \citep{maa22yoo} proposes two separate methods, NodeSam and SubMix, to generate synthetic samples by heuristic structure modification. Besides, GLA \citep{gla22yue} generates the latent representations as the augmented graphs during the training phase. Subsequently, GMixup \citep{gmixup22han} and FGWMixup \citep{fgwmixup23ma} interpolate graphs or features linearly to mix normal and anomalous samples for producing novel samples. Nevertheless, they fail to produce semantic-preserving samples when dealing with imbalanced data with limited supervision, resulting in unsatisfactory performance. 

In contrast, FracAug diverges by leveraging the fractional power of adjacency matrices, a mathematically grounded operation that preserves semantics within graphs while introducing multi-scale structural variations. The incorporation of weighted distance-aware margin loss further enables FracAug to adapt to the imbalanced scenario. Furthermore, its pseudo-labeling mechanism explicitly addresses the limited supervision constraint. As a plug-in module, FracAug overcomes above limitations in GNNs and graph-level augmentation methods without modifying GNN architectures, enabling given GNN models to learn discriminative features for GAD tasks even with sparse labels.

\section{Preliminaries}
\label{sec:preliminaries}
{\bf Notation.} 
Let $G = (\vect{A},\vect{X})$ denote an undirected graph with $n$ nodes and $m$ edges, where $\vect{A}\in \mathbb{R}^{n\times n}$ is the adjacency matrix and $\vect{X} \in \mathbb{R}^{n\times F}$ is the node feature matrix. $\vect{A}_{ij}=1$ if an edge exists between node $i$ and $j$, and $\vect{A}_{ij}=0$ otherwise. $\vect{D}$ is the diagonal degree matrix of $\vect{A}$, and the normalized adjacency matrix can be defined as $\tilde{\vect{A}}=\vect{D}^{-\frac{1}{2}}\vect{A}\vect{D}^{-\frac{1}{2}}$ correspondingly. For a given matrix $\vect{M}$, $\vect{M}^\alpha$ stands for the $\alpha$-th power of matrix $\vect{M}$, where $\alpha\geq 0$. When $\vect{M}$ can be eigendecomposed, $\vect{M}^\alpha=\vect{U}\vect{\Lambda}^\alpha\vect{V}$, where $\vect{U}, \vect{V}\in \mathbb{C}^{n\times n}$ are unitary matrices and $\vect{\Lambda}\in\mathbb{R}^{n\times n}$ is a diagonal matrix composed of the eigenvalues of $\vect{M}$. 

{\bf Continuous Semantic Space.} Given a graph $G$ with adjacency matrix $\vect{A}$ and a graph signal $\vect{x}\in \mathbb{R}^F$, the semantic space of $G$ is defined as a subspace $\mathcal{S}\subseteq \mathbb{R}^F$ generated by the set of vectors obtained through the application of powers of $\vect{A}$ to $\vect{x}$. Unlike previous approaches that rely on discrete semantics constrained by integer powers, i.e., $\{\vect{A}^t\vect{x} | t\in \mathbb{N}\}$, our continuous semantic space formulation, $\mathcal{S}=span\{\vect{A}^t\vect{x}|t\in \mathbb{N}\}$, captures the underlying continuous semantic manifold of the graph, which enables us to synthesize novel semantic-preserving graph instances, shown in Section \ref{sec:method}.

{\bf Graph-level Anomaly Detection.} In this work, we focus on enhancing GAD performance under limited supervision. Given a training set with $k$ labeled samples, $\mathcal{D}_{train}=\{(G_1, y_1), (G_2, y_2), ..., (G_k, y_k)\}$, the goal of GAD is to train a model that classifies unseen graphs as normal or anomalous. In real deployment, there are two main challenges in GAD. Firstly, $\mathcal{D}_{all}=\mathcal{D}_{train}\cup\mathcal{D}_{val}\cup\mathcal{D}_{test}$ contains $N_0$ normal graphs and $N_1$ anomalous graphs, where $N_0\gg N_1$, leading to severe imbalanced problem. Secondly, only limited labeled graphs are accessible during training, i.e., $k\ll N_0+N_1$, resulting in the limited supervision issue. Therefore, the key to enhancing the ability of GNNs on real-world GAD tasks is to address these two challenges simultaneously. 

{\bf Graph-level Augmentation.} Graph-level augmentation has been proven effective in improving the performance of GNNs on graph-level tasks. Graph generation and pseudo-labeling are the most common ways to conduct graph-level augmentation: 
\begin{itemize} [topsep=0.5mm, partopsep=0pt, itemsep=0pt, leftmargin=10pt]
    \item Graph generation: This strategy maps the graph $G\in \mathcal{D}_{train}$ to a new graph $G'$, i.e., $(G,y)\mapsto (G',y)$. The generated graph should have a semantic meaning similar to that of the original $G$. 
    \item Pseudo-labeling: This approach leverages a GNN trained on $\mathcal{D}_{train}$ to classify samples from $\mathcal{D}_{val}\cup\mathcal{D}_{test}$ and assign pseudo-labels to samples with high confidence under a certain criterion. 
\end{itemize}

By combining graph generation and pseudo-labeling techniques while tackling the imbalanced issue, FracAug effectively boosts GNN performance for GAD under limited supervision.

\section{Method}
\label{sec:method}

\subsection{Overview}
\label{subsec:overview}
\begin{figure}[t]
\centering

\vspace{-5mm}
\includegraphics[height=42mm]{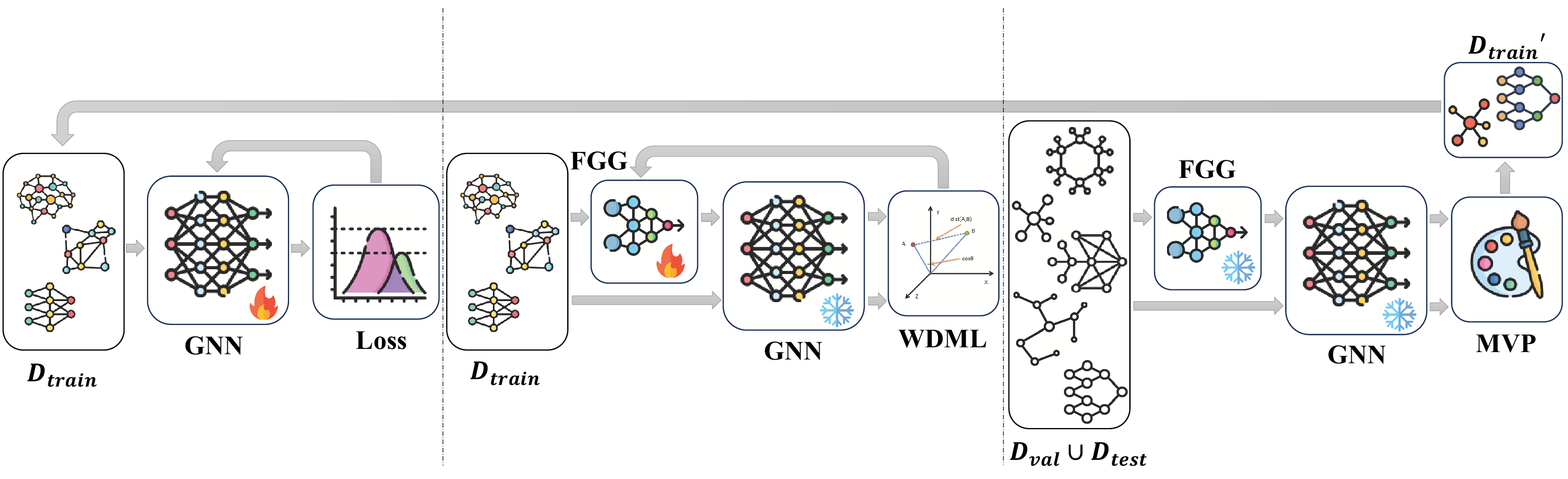}
\vspace{-3mm}
\caption{Overview of FracAug.}
\label{fig:framework}
\vspace{-5mm}
\end{figure}

Our proposed FracAug consists of three key components: (1) {\bf Fractional Graph Generator (FGG)} in Section \ref{subsec:fractionalgraphgenerator} captures the inherent semantics of graphs, enabling the synthesis of fractional variants that maintain semantic consistency with originals, as we demonstrate theoretically. (2) {\bf Weighted Distance-Aware Margin Loss (WDML)} in Section \ref{subsec:weightedistanceawaremarginloss} addresses data imbalance to guide FGG, employing distance-based margins to position synthetic graphs near original counterparts while ensuring distinctiveness. (3) {\bf Mutual Verification Pseudo-Labeler (MVP)} in Section \ref{subsec:mutualverificationpseudolabeler} minimizes pseudo-labeling errors through mutual verification of predictions from original and synthetic graphs, facilitating reliable and iterative training set expansion. 

Figure \ref{fig:framework} illustrates the pipeline of our FracAug. Initially, we warm up a given GNN to establish a preliminary semantic understanding of the GAD task. Then, we freeze the GNN parameters and utilize its outputs to train the FGG with WDML. The trained FGG then generates fractional graph variants, and the GNN predicts on both original and synthetic graphs to pseudo-label data within the validation and test sets using MVP, which are subsequently incorporated into the original training set. Finally, we train the GNN using the new training set and continue the above process until both the GNN and our FracAug framework reach reasonable capability.

\subsection{Fractional Graph Generator}
\label{subsec:fractionalgraphgenerator}

{\bf Flexible Eigengraph Combinations.} The fractional power of the adjacency matrix, $\vect{A}^\alpha$, where $\alpha \geq 0$, serves as the mathematical foundation of our framework due to its unique properties. Unlike integer powers of $\vect{A}$, which only capture discrete-step neighborhood aggregations, fractional powers enable continuous interpolation of graph structures, providing fine-grained control over topological variations. Crucially, $\vect{A}^\alpha$ can be expressed as a combination of eigengraphs derived from eigendecompositions. For an undirected graph $G$ with symmetric adjacency matrix $\vect{A}$, we can decompose $\vect{A}^\alpha$ as: 
\begin{equation*}
    \vect{A}^\alpha=\vect{U}\vect{\Lambda}^\alpha\vect{U}^T=\sum_{i=1}^n\lambda_i^\alpha \vect{u}_i\vect{u}_i^T, 
\end{equation*}
where $\vect{\Lambda}$ is the diagonal eigenvalue matrix containing $\{\lambda_i\}_{i=1}^n$ in a descending order, $\vect{U}$ is the eigenvector matrix formed by $\{\vect{u}_i\}_{i=1}^n$, and $\vect{u}_i\vect{u}_i^T$ is the $i$-th eigengraph. It reveals two key advantages:
\begin{itemize} [topsep=0.5mm, partopsep=0pt, itemsep=0pt, leftmargin=10pt]
    \item Multi-scale Structure Adaptation: Fractional powers enable tunable control over spectral components via $\alpha$, where lower values $(\alpha < 1)$ emphasize homophilic graph signals (low-frequency eigengraphs), while higher values $(\alpha > 1 )$ accentuate heterophilic graph signals (high-frequency eigengraphs) \citep{negativedepth23yan}. This adaptive reweighting preserves the hierarchical topology while generating augmented graphs, signaling structural anomalies for detection. 
    \item Semantic-preserving Combination: By combining eigengraphs, FracAug preserves semantic-critical structures (targeting spectral deviations linked to anomalies \citep{rqgnn24dong}), ensuring that generated graphs retain the original semantics.
\end{itemize}

{\bf Semantic Preservation.} 
Prior studies, such as GIN \citep{gin19xu}, rely on integer powers of adjacency matrices, limiting them to discrete semantic preservation. In contrast, we prove that for any $\alpha \geq 0$, $\vect{A}^\alpha\vect{x}$ resides in the original semantic space. Moreover, we further derive a theoretical boundary to quantify differences between the original and fractional graphs, detailed in Appendix \ref{app:proof}.
\begin{theorem} \label{thm:approximationfractional}
    Given a polynomial function $p(\cdot;\vect{\theta})$ parameterized by $\vect{\theta}$, for any $\alpha\ge 0$, there exists $\vect{\theta}^*$ such that $\vect{A}^\alpha\approx p(\vect{A};\vect{\theta}^*)=\sum_{t=0}^{T}\vect{\theta}_t^*\vect{A}^t, T\in\mathbb{N}$. With proper parameter $\vect{\theta}^*$, the difference of them is bounded by $\beta e^{-\gamma T}$, where $\beta, \gamma>0$ depend on the eigenvalues of $\vect{A}$. Since $\vect{A}^\alpha$ can be represented as a polynomial combination of $\{\vect{A}^t\}_{t\in \mathbb{N}}$, $\vect{A}^\alpha\vect{x}$ lies in $\mathcal{S}$ of the original graph as $T\rightarrow\infty$. 
\end{theorem}

Theorem \ref{thm:approximationfractional} ensures that fractional graphs preserve the original semantic space while encoding multi-scale semantics. The continuous parameter $\alpha$ spans all possible semantic variations within this space, enabling rich and comprehensive augmentation. Besides, previous approaches such as MAA \citep{maa22yoo} leverage heuristic perturbation techniques for generating synthetic graphs, which may result in useful substitutes near the semantic space of the original graph. Theorem \ref{thm:approximationperturbation} formally bridges these perturbation-based approaches with our fractional graph augmentation, revealing their shared theoretical foundations. The complete proof is provided in Appendix \ref{app:proof}.

\begin{theorem} \label{thm:approximationperturbation}
    Let the structural perturbation on a graph be a perturbation matrix $\vect{P}$ added to the original graph, so that any graph generated by a structural perturbation method can be expressed as $\vect{A}+\vect{P}$. Then, we can derive $||\vect{A}^\alpha - (\vect{A}+\vect{P})||\leq c||\vect{P}||+\max_i|\lambda_i-\lambda_i^\alpha|$, where $c$ depends on $\alpha$ and the spectral gap of $\vect{P}$, and $\lambda_i$ is the $i$-th eigenvalue of $\vect{A} + \vect{P}$. Thus, by choosing an appropriate $\alpha$, $\vect{A}^\alpha$ can approximate any graph generated by structural perturbation methods. 
\end{theorem}

Based on Theorem \ref{thm:approximationperturbation}, we observe that for a suitably chosen $\alpha$, the fractional graph can approximate any sample generated by perturbation-based framework, demonstrating its generalization capability. 

{\bf Fractional Graph Generation.} Building on the above analysis, we conclude that fractional graphs offer powerful augmentation capabilities. However, directly deriving the fractional power of the adjacency matrix can be computationally prohibitive and may yield invalid results for non-semi-definite adjacency matrix. Thus, a transformation function $h(\cdot)$ is applied to the adjacency matrix to ensure valid fractional powers while preserving structural integrity \citep{negativedepth23yan}. Specifically, instead of adding self-loops before normalization of the adjacency matrix, we introduce them after normalization and rescale the matrix, so the resulting adjacency matrix can be defined as:
\begin{equation*}
    \hat{\vect{A}}=h(\tilde{\vect{A}})=\frac{1}{2}(\vect{I}+\tilde{\vect{A}}), 
\end{equation*}
which is still a normalized adjacency matrix. Since all eigenvalues of $\tilde{\vect{A}}$ lie within $[-1, 1]$, the corresponding eigenvalues of $\hat{\vect{A}}$ fall within $[0, 1]$. Therefore, $h(\cdot)$ transforms $\vect{A}$ into a positive semi-definite matrix $\hat{\vect{A}}$, which allows the design of FGG. 

Moreover, to mitigate computational costs for large graphs, we precompute the eigendecomposition (EVD) using the Arnoldi method \citep{arnoldi98lehoucq}, retaining only the top-$k_l$ largest and top-$k_s$ smallest eigenpairs. Denote $\hat{\vect{\Lambda}}_{k_l}, \hat{\vect{\Lambda}}_{k_s}$ as diagonal matrices of the top-$k_l$ largest and top-$k_s$ smallest eigenvalues, $\vect{U}_{k_l}=\vect{U}[:, 0:k_l], \vect{U}_{k_s}=\vect{U}[:, n-{k_s}:n]$ as the corresponding matrices of eigenvectors, and the generated graph of $G(\vect{A}, \vect{X})$ as $G'(\vect{A}', \vect{X})$, FGG can be formulated as: 
\begin{equation*}
\begin{aligned}
    g(\vect{A}, k, H)&=\sum_{h=1}^H\omega_h\vect{U}_k\hat{\vect{\Lambda}}_k^{\alpha_h}\vect{U}_k^T, \\
    \vect{A}'=\text{FGG}(\vect{A}, k_l, k_s, H_l, H_s)&=\omega g(\vect{A}, k_l, H_l)+(1-\omega)g(\vect{A}, k_s, H_s), 
\end{aligned}
\end{equation*}
where $\sum_{h=1}^H\omega_h=1$, $\omega$ are learnable coefficients, and $\alpha_h$ is $h$-th learnable fractional power of the matrix. By combining multiple fractional graphs with tunable weights, our generated graphs can capture comprehensive information while preserving semantics. Although the anomalous properties are well-preserved in graphs from FGG based on previous analysis, inherent data imbalance risks biasing FGG training. To counteract this, in Section \ref{subsec:weightedistanceawaremarginloss}, we design WDML to guide FGG training.

\subsection{Weighted Distance-Aware Margin Loss}
\label{subsec:weightedistanceawaremarginloss}
{\bf Revisiting Margin Loss.} To better separate the semantic spaces of different graphs and enable FGG to generate high-quality fractional graphs robust to class imbalance, we introduce a novel margin loss function. Before illustrating the details of WDML, we first reexamine representative margin losses, with comprehensive empirical validation provided in Appendix \ref{app:lossfunctioncomoparson}. Formally, margin loss based on cross-entropy can be defined as: 
\begin{equation} \label{equ:margincrossentropy}
    L=-\frac{1}{N}\sum_{i=1}^N\log \frac{e^{\vect{s}_{\vect{y}_i}-m}}{e^{\vect{s}_{\vect{y}_i}-m}+\sum_{j=1, j\neq \vect{y}_i}^Ce^{\vect{s}_{j}}}, 
\end{equation}
where $N$ is the number of the samples, $\vect{s}$ represents the normalized logits predicted by a given GNN, $\vect{y}_i$ is the ground truth label of $i$-th sample, $m$ is the margin that determines the decision boundary, and $C$ is the number of classes. 

\begin{figure}[t]
\centering
 \vspace{-2mm}
  \begin{small}
    \begin{tabular}{cccc}
        \hspace{-6mm}
        \includegraphics[height=20mm]{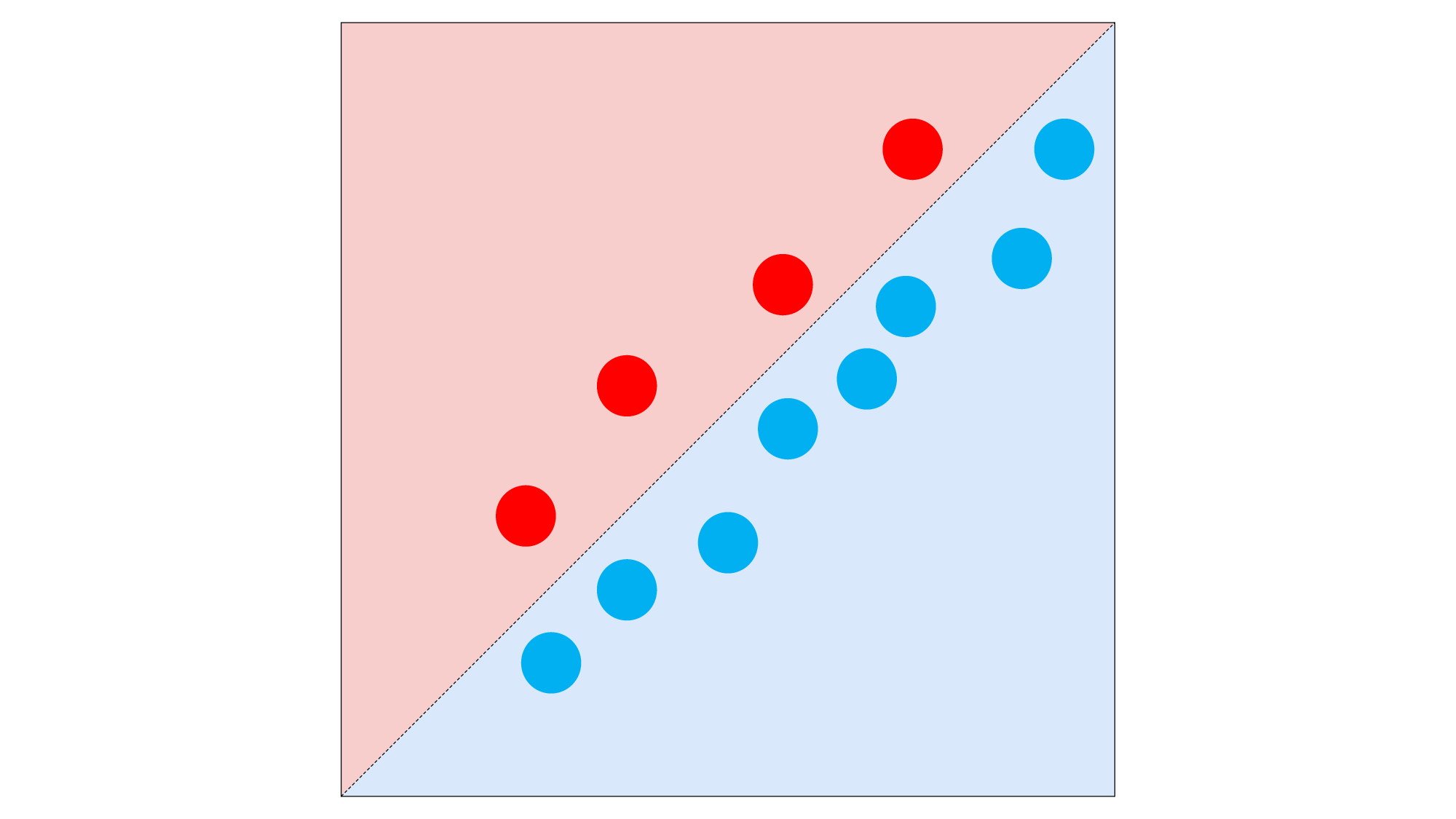} &
        \hspace{-6mm}
        \includegraphics[height=20mm]{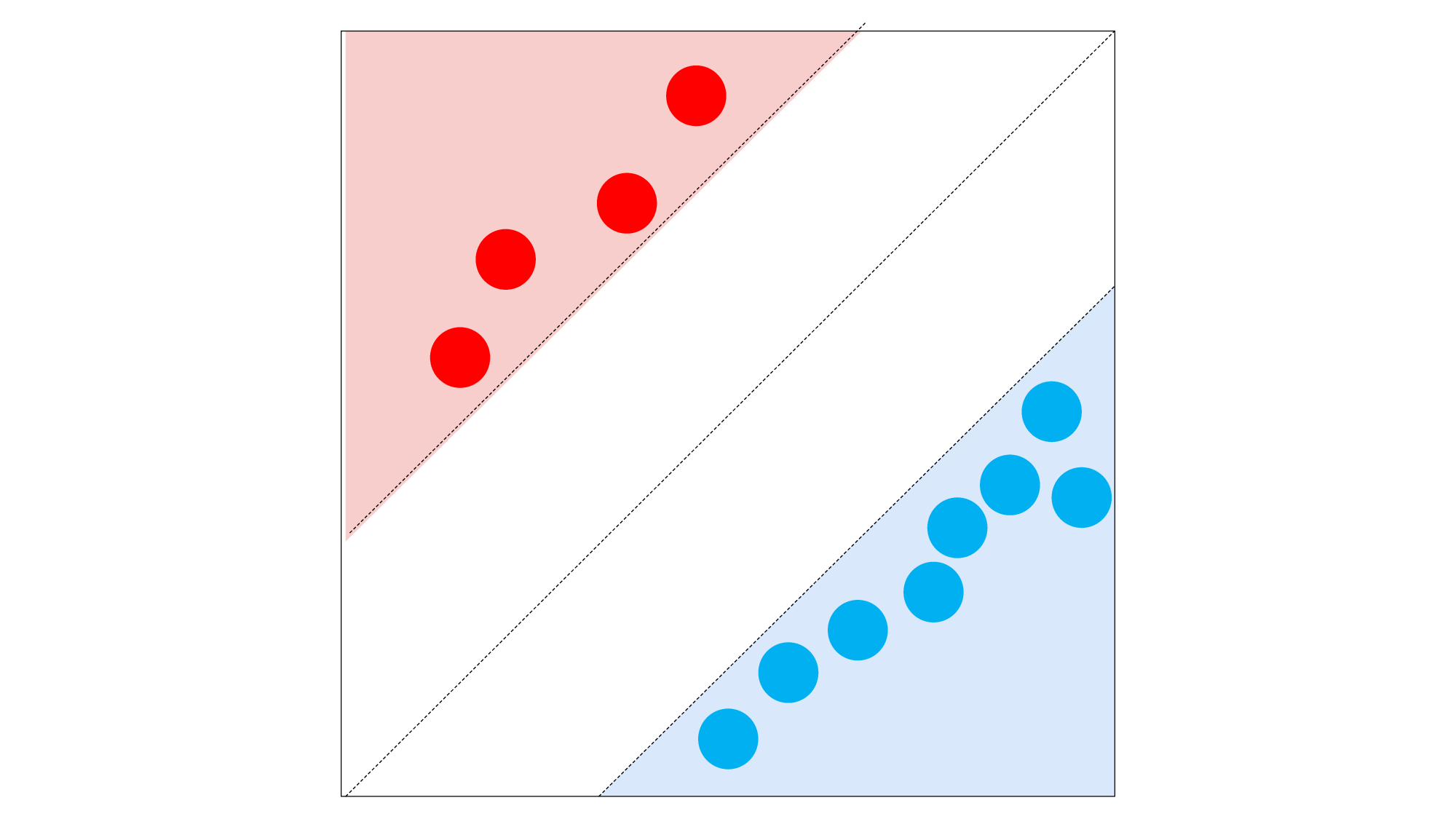} &
        \hspace{-6mm}
        \includegraphics[height=20mm]{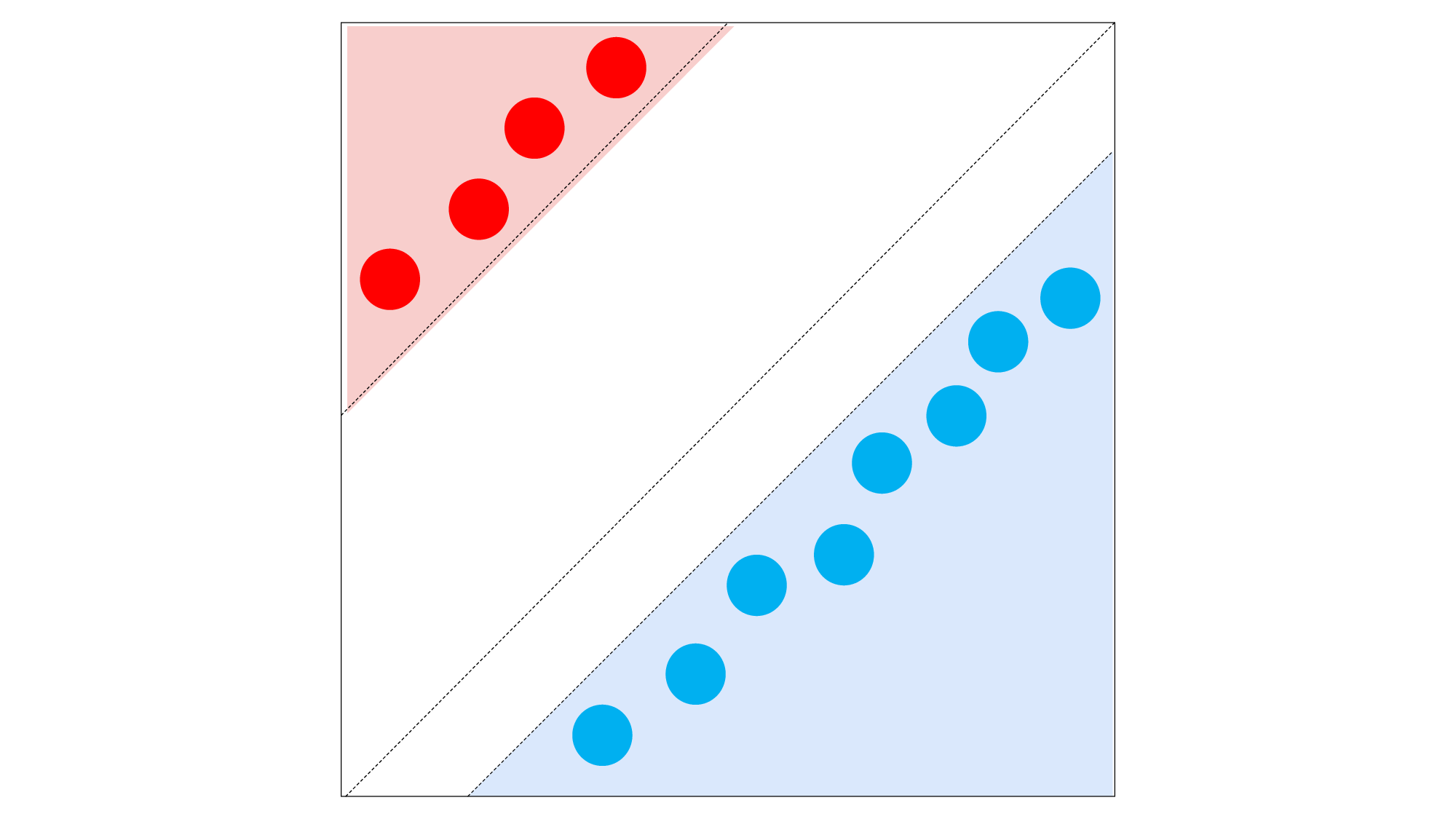} &
        \hspace{-6mm}
        \includegraphics[height=20mm]{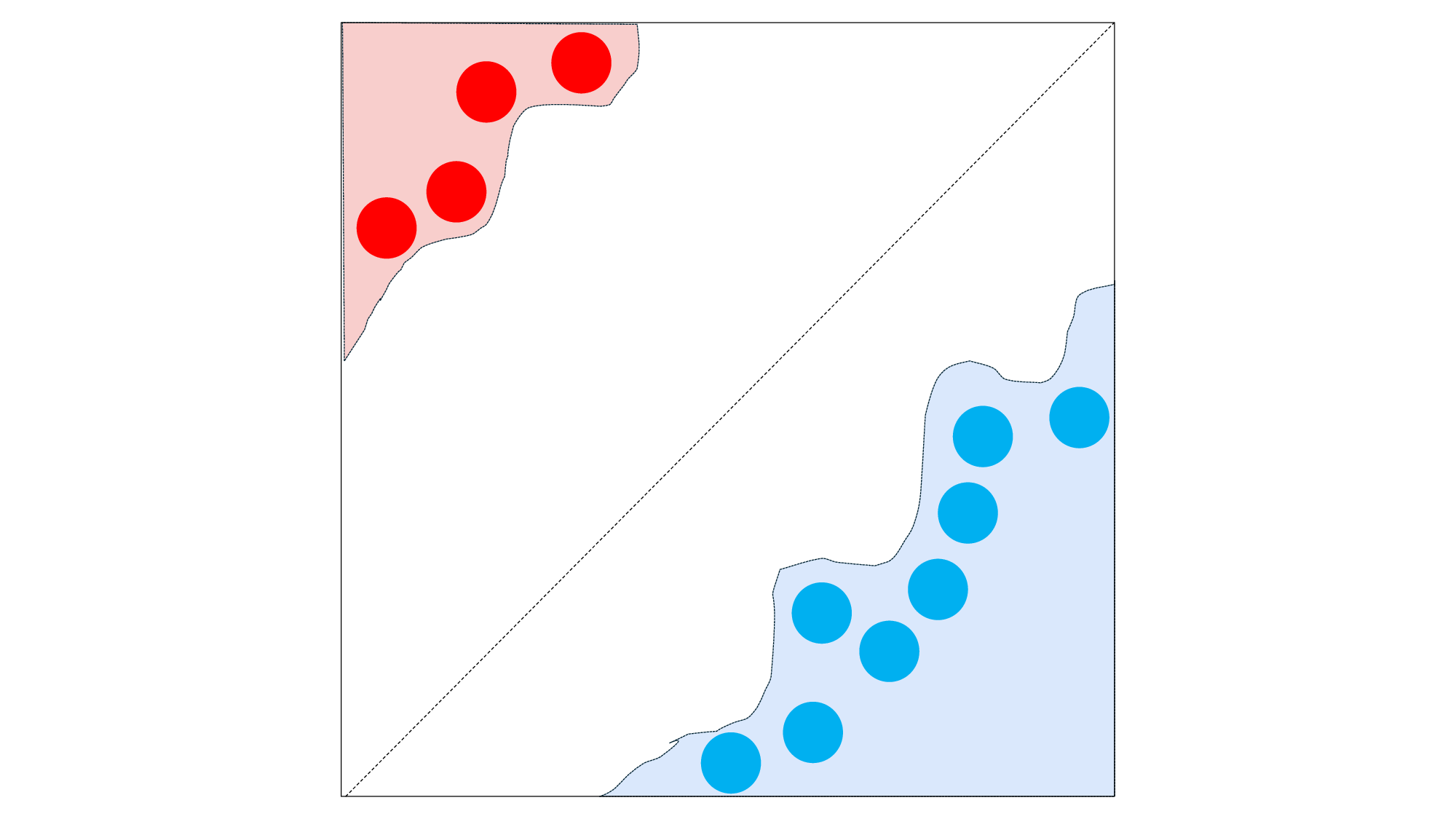} \\ 
        \hspace{-6mm}
        (a) Softmax & 
        \hspace{-6mm}
        (b) LMCL & 
        \hspace{-6mm}
        (c) LDAM & 
        \hspace{-6mm}
        (d) WDML \\ 
    \end{tabular}
    \caption{Decision boundaries of different margin losses.}
    \label{fig:decisionboundary}
    \vspace{-4mm}
  \end{small}
\end{figure}

As shown in Figure \ref{fig:decisionboundary} (a), when setting margin $m$ to $0$, the margin loss is degraded to a cross-entropy loss, which lacks explicit mechanisms to separate classes in complex scenarios. Another margin loss is LMCL \citep{lmcl18wang} with $m$ as a hyperparameter. Figure \ref{fig:decisionboundary} (b) describes the result of setting $m>0$, enforcing better inter-class separation. However, this uniform margin shifts the decision boundaries of different classes by the same value, which fails to detect the unique class-specific properties. Afterward, LDAM \citep{ldam19cao} in Figure \ref{fig:decisionboundary} (c) tackles the issues by setting class-specific margin $m_c$ for $c$-th class so that the decision boundaries can accommodate scenarios where classes require distinct margins. Existing margin losses typically employ fixed margins, which prove suboptimal for GAD where sample-specific semantic variations exist. To address this, we propose WDML, which assigns dynamic margins based on the intrinsic distance of each synthetic sample and its original graph with a weight according to its class. Figure \ref{fig:decisionboundary} (d) describes the adaptive decision boundary of WDML.

{\bf Margin Loss Based on Sample-Specific Distance.} For the $i$-th training graph $G_i$ and its counterpart $G_i'$ generated by FGG, we extract graph-level embeddings $\vect{o}_i$ and $\vect{o}'_i$ via a given GNN. Then, our distance-aware margin can be defined as: 
\begin{equation} \label{equ:margin}
    m_i=\frac{1-\cos(\vect{o}_i, \vect{o}_i')}{2}, 
\end{equation}
where $\cos$ represents cosine similarity. Substituting $m$ in Equation \ref{equ:margincrossentropy} with the sample-specific margin $m_i$ yields a distance-aware margin loss. By computing angular distances in Equation \ref{equ:margin}, this loss shifts the semantic space away from decision boundaries by a margin $m_i$, ensuring generated samples retain the original label with high confidence. To further address class imbalance, WDML incorporates weights based on class frequency: 
\begin{equation*}
    L_{\text{WDML}}=-\sum_{i=1}^N\frac{1}{N_{\vect{y}_i}}\log \frac{e^{\vect{s}_{\vect{y}_i}-m_i}}{e^{\vect{s}_{\vect{y}_i}-m_i}+\sum_{j=1, j\neq \vect{y}_i}^Ce^{\vect{s}_{j}}}, 
\end{equation*}
where $N_{\vect{y}_i}$ is the number of samples in class $\vect{y}_i$. With the assistance of WDML, FGG can generate fractional graphs effectively without being biased by the imbalanced distribution of labels. To further boost the performance by data augmentation, we design MVP to combine graph generation and pseudo-labeling techniques, whose details will be elaborated in Section \ref{subsec:mutualverificationpseudolabeler}.

\subsection{Mutual Verification Pseudo-Labeler}
\label{subsec:mutualverificationpseudolabeler}
{\bf Insight on Mutual Verification.} Prior pseudo-labeling methods for related tasks, such as ConsisGAD \citep{consisgad24chen}, only rely on confidences from original samples, prone to high errors under low supervision \citep{spacegnn24dong}. Therefore, we first investigate how mutual verification mitigates the error rates compared to single-view methods, theoretically. The proof is detailed in the Appendix \ref{app:proof}. 
\begin{proposition} \label{pro:reductionfactor}
    For a given GNN, assume its prediction error rates for original graphs and corresponding fractional graphs are both $\delta$. The correlation coefficient between the errors is denoted as $\rho$. Then, when mutual verification is used, compared to single-view methods, the reduction factor of the error rate and its variance can be up to $\delta+\rho\delta(1-\delta)$ and $\rho$, respectively. 
\end{proposition}
Proposition \ref{pro:reductionfactor} demonstrates that the mutual verification mechanism leverages semantic consistency between original graphs and their fractional counterparts to enhance pseudo-labeling reliability. Building on this, we design MVP based on the agreement between predicted labels for the original and synthetic samples, as detailed below. 

{\bf High-Quality Pseudo-Label Prediction.} Based on the above analysis, MVP assigns a pseudo-label $\hat{y}_i$ to the $i$-th sample in the validation or test set if and only if: 
\begin{equation*}
    \hat{y}_i=\left\{ \begin{aligned}
        0,\ \ \ p_i\leq\tau_n \land p_i'\leq \tau_n, \\
        1,\ \ \ p_i\geq\tau_a \land p_i'\geq \tau_a, 
    \end{aligned}
    \right. 
\end{equation*}
where $p_i, p_i'$ represent the anomaly probabilities of the $i$-th original sample and its fractional counterpart, respectively, and $\tau_n, \tau_a$ denote the confidence thresholds of a sample being normal/anomalous. For any given GNN, we iteratively incorporate high-confidence pseudo-labeled samples from the validation and test sets into the training set, further mitigating the limited supervision issue. 

The core innovation of our mutual verification framework lies in leveraging the semantic consistency between original and fractional graphs to generate high-confidence pseudo-labels. This mechanism addresses the scarcity of labeled anomalies by iteratively expanding the training set with reliable samples, guided by theoretical guarantees of robustness.

In summary, our proposed FracAug combines FGG, WDML, and MVP to generate fractional graphs and pseudo-label samples to boost the performance of GNNs on GAD tasks under limited supervision. The experiments in Section \ref{sec:experiments} further validate our theoretical analysis in Section \ref{sec:method}.

\section{Experiments}
\label{sec:experiments}

\begin{table}[t]
\caption{Average AUROC, AUPRC, and F1-score on 12 datasets with multiple runs, using graph classification models as baselines, where the white columns represent vanilla models and the gray ones represent models augmented by FracAug. }
\small
\centering
\scalebox{0.91}{
\setlength\tabcolsep{1pt}
\label{tab:gcperformance}
\begin{tabular}{cc|c
>{\columncolor[HTML]{dcdcdc}}c| c
>{\columncolor[HTML]{dcdcdc}}c| c
>{\columncolor[HTML]{dcdcdc}}c| c
>{\columncolor[HTML]{dcdcdc}}c| c
>{\columncolor[HTML]{dcdcdc}}c| c
>{\columncolor[HTML]{dcdcdc}}c }
\hline \hline
Datasets                         & Metrics  & GCN             & \multicolumn{1}{l|}{\cellcolor[HTML]{ffffff}{\scriptsize +FA}} & SAGE            & \multicolumn{1}{l|}{\cellcolor[HTML]{ffffff}{\scriptsize +FA}} & GAT             & \multicolumn{1}{l|}{\cellcolor[HTML]{ffffff}{\scriptsize +FA}} & GIN    & \multicolumn{1}{l|}{\cellcolor[HTML]{ffffff}{\scriptsize +FA}} & LRGNN                          & \multicolumn{1}{l|}{\cellcolor[HTML]{ffffff}{\scriptsize +FA}} & GRDL            & \multicolumn{1}{l}{\cellcolor[HTML]{ffffff}{\scriptsize +FA}} \\ \hline 
                                 \multirow{3}{*}{MCF-7}          & AUROC    & 0.5753 & \textbf{0.5840} & 0.5801          & \textbf{0.5930} & 0.5885          & \textbf{0.6058} & 0.5867 & \textbf{0.5976} & 0.5467          & \textbf{0.6117} & 0.5867          & \textbf{0.6197} \\
                                & AUPRC    & 0.3035 & \textbf{0.3073} & 0.3316          & \textbf{0.3340} & 0.3678          & \textbf{0.3814} & 0.2830 & \textbf{0.2971} & 0.2951          & \textbf{0.3309} & 0.3038          & \textbf{0.3469} \\
                                & F1-score & 0.4982 & \textbf{0.5074} & 0.4798          & \textbf{0.4963} & 0.4573          & \textbf{0.4687} & 0.5366 & \textbf{0.5421} & 0.4660          & \textbf{0.5290} & 0.5147          & \textbf{0.5254} \\ \hline
\multirow{3}{*}{MOLT-4}         & AUROC    & 0.5531 & \textbf{0.5650} & 0.5326          & \textbf{0.5727} & 0.5403          & \textbf{0.5721} & 0.5733 & \textbf{0.5854} & 0.5495          & \textbf{0.5851} & 0.5858          & \textbf{0.5924} \\
                                & AUPRC    & 0.3183 & \textbf{0.3217} & 0.1907          & \textbf{0.2602} & 0.3456          & \textbf{0.3667} & 0.2830 & \textbf{0.3001} & 0.3047          & \textbf{0.3175} & 0.3018          & \textbf{0.3101} \\
                                & F1-score & 0.4501 & \textbf{0.4626} & 0.5166          & \textbf{0.5277} & 0.4096          & \textbf{0.4313} & 0.5072 & \textbf{0.5103} & 0.4570          & \textbf{0.4939} & 0.5091          & \textbf{0.5117} \\ \hline
\multirow{3}{*}{PC-3}           & AUROC    & 0.5697 & \textbf{0.5863} & 0.5986          & \textbf{0.6119} & 0.5707          & \textbf{0.5865} & 0.5969 & \textbf{0.6119} & 0.5690          & \textbf{0.6102} & 0.6044          & \textbf{0.6202} \\
                                & AUPRC    & 0.2740 & \textbf{0.2837} & 0.3154          & \textbf{0.3248} & 0.3562          & \textbf{0.3626} & 0.2797 & \textbf{0.2893} & 0.2320          & \textbf{0.3038} & 0.3381          & \textbf{0.3459} \\
                                & F1-score & 0.4745 & \textbf{0.4876} & 0.4751          & \textbf{0.4841} & 0.4036          & \textbf{0.4166} & 0.5063 & \textbf{0.5205} & \textbf{0.5103} & 0.5024          & 0.4615          & \textbf{0.4750} \\ \hline
\multirow{3}{*}{SW-620}         & AUROC    & 0.5662 & \textbf{0.5839} & 0.5800          & \textbf{0.5968} & 0.5633          & \textbf{0.5870} & 0.5938 & \textbf{0.6004} & 0.5758          & \textbf{0.5946} & 0.6005          & \textbf{0.6046} \\
                                & AUPRC    & 0.3134 & \textbf{0.3229} & \textbf{0.3401} & 0.3260          & 0.2481          & \textbf{0.2587} & 0.2776 & \textbf{0.2813} & \textbf{0.3506} & 0.3386          & \textbf{0.2908} & 0.2901          \\
                                & F1-score & 0.4406 & \textbf{0.4541} & 0.4339          & \textbf{0.4678} & 0.4923          & \textbf{0.5187} & 0.5090 & \textbf{0.5155} & 0.4190          & \textbf{0.4536} & 0.5059          & \textbf{0.5135} \\ \hline
\multirow{3}{*}{NCI-H23}        & AUROC    & 0.5811 & \textbf{0.5864} & 0.5765          & \textbf{0.6105} & 0.5777          & \textbf{0.6084} & 0.5897 & \textbf{0.5968} & 0.6002          & \textbf{0.6315} & 0.6161          & \textbf{0.6271} \\
                                & AUPRC    & 0.2777 & \textbf{0.2777} & 0.3197          & \textbf{0.3207} & \textbf{0.2966} & 0.2945          & 0.2566 & \textbf{0.2659} & 0.2830          & \textbf{0.3205} & 0.3034          & \textbf{0.3069} \\
                                & F1-score & 0.4751 & \textbf{0.4819} & 0.4333          & \textbf{0.4740} & 0.4548          & \textbf{0.4961} & 0.5059 & \textbf{0.5073} & 0.4987          & \textbf{0.5055} & 0.4983          & \textbf{0.5112} \\ \hline
\multirow{3}{*}{OVCAR-8}        & AUROC    & 0.5692 & \textbf{0.5809} & 0.5763          & \textbf{0.5836} & 0.5396          & \textbf{0.5784} & 0.5935 & \textbf{0.5963} & 0.5628          & \textbf{0.5984} & 0.6230          & \textbf{0.6311} \\
                                & AUPRC    & 0.3240 & \textbf{0.3263} & 0.3391          & \textbf{0.3423} & 0.2010          & \textbf{0.2401} & 0.2573 & \textbf{0.2612} & 0.3106          & \textbf{0.3290} & 0.3042          & \textbf{0.3129} \\
                                & F1-score & 0.4216 & \textbf{0.4334} & 0.4163          & \textbf{0.4221} & 0.4855          & \textbf{0.5060} & 0.5118 & \textbf{0.5123} & 0.4260          & \textbf{0.4518} & 0.5087          & \textbf{0.5119} \\ \hline
\multirow{3}{*}{P388}           & AUROC    & 0.5171 & \textbf{0.5896} & 0.5820          & \textbf{0.6277} & 0.4964          & \textbf{0.5758} & 0.5565 & \textbf{0.5913} & 0.5546          & \textbf{0.6316} & 0.5500          & \textbf{0.5852} \\
                                & AUPRC    & 0.3488 & \textbf{0.3926} & \textbf{0.3569} & 0.3540          & 0.2045          & \textbf{0.2134} & 0.2850 & \textbf{0.3309} & 0.2880          & \textbf{0.2953} & 0.2318          & \textbf{0.2859} \\
                                & F1-score & 0.3428 & \textbf{0.3886} & 0.4138          & \textbf{0.4741} & 0.4478          & \textbf{0.5481} & 0.4468 & \textbf{0.4491} & 0.4430          & \textbf{0.5496} & 0.4808          & \textbf{0.4814} \\ \hline
\multirow{3}{*}{SF-295}         & AUROC    & 0.5730 & \textbf{0.5813} & 0.5858          & \textbf{0.6057} & 0.5960          & \textbf{0.6171} & 0.5844 & \textbf{0.6076} & 0.5903          & \textbf{0.6185} & 0.6156          & \textbf{0.6349} \\
                                & AUPRC    & 0.3290 & \textbf{0.3308} & 0.3161          & \textbf{0.3222} & 0.2652          & \textbf{0.2708} & 0.2766 & \textbf{0.2832} & \textbf{0.3000} & 0.2972          & 0.2796          & \textbf{0.3115} \\
                                & F1-score & 0.4199 & \textbf{0.4279} & 0.4463          & \textbf{0.4652} & 0.5065          & \textbf{0.5389} & 0.4803 & \textbf{0.5047} & 0.4669          & \textbf{0.5068} & \textbf{0.5221} & 0.5173          \\ \hline
\multirow{3}{*}{SN12C}          & AUROC    & 0.5624 & \textbf{0.5818} & 0.5705          & \textbf{0.6030} & 0.5863          & \textbf{0.6020} & 0.5995 & \textbf{0.6079} & 0.5973          & \textbf{0.6104} & 0.6061          & \textbf{0.6211} \\
                                & AUPRC    & 0.2812 & \textbf{0.2981} & 0.2859          & \textbf{0.3133} & 0.2468          & \textbf{0.2585} & 0.2696 & \textbf{0.2746} & 0.2729          & \textbf{0.2888} & 0.2803          & \textbf{0.2875} \\
                                & F1-score & 0.4463 & \textbf{0.4546} & 0.4514          & \textbf{0.4670} & 0.5058          & \textbf{0.5191} & 0.5030 & \textbf{0.5110} & 0.4978          & \textbf{0.5012} & 0.5026          & \textbf{0.5183} \\ \hline
\multirow{3}{*}{UACC257}        & AUROC    & 0.5660 & \textbf{0.5831} & 0.6006          & \textbf{0.6132} & 0.5890          & \textbf{0.6174} & 0.5877 & \textbf{0.6015} & 0.6020          & \textbf{0.6189} & 0.6155          & \textbf{0.6340} \\
                                & AUPRC    & 0.3334 & \textbf{0.3509} & \textbf{0.3360} & 0.3337          & \textbf{0.3493} & 0.3389          & 0.2480 & \textbf{0.2598} & 0.3047          & \textbf{0.3215} & 0.2942          & \textbf{0.3051} \\
                                & F1-score & 0.3921 & \textbf{0.3954} & 0.4289          & \textbf{0.4456} & 0.4031          & \textbf{0.4455} & 0.4906 & \textbf{0.4983} & 0.4585          & \textbf{0.4631} & 0.4843          & \textbf{0.4990} \\ \hline
\multirow{3}{*}{PROTEINS\_full} & AUROC    & 0.6186 & \textbf{0.6259} & 0.5942          & \textbf{0.6310} & 0.6157          & \textbf{0.6836} & 0.5799 & \textbf{0.6174} & 0.6434          & \textbf{0.6503} & 0.5895          & \textbf{0.5987} \\
                                & AUPRC    & 0.6325 & \textbf{0.6404} & 0.6086          & \textbf{0.6516} & 0.6350          & \textbf{0.7005} & 0.6259 & \textbf{0.6358} & 0.6603          & \textbf{0.6722} & 0.6015          & \textbf{0.6077} \\
                                & F1-score & 0.6199 & \textbf{0.6273} & 0.5909          & \textbf{0.6289} & 0.6158          & \textbf{0.6859} & 0.5679 & \textbf{0.6175} & 0.6431          & \textbf{0.6469} & 0.5856          & \textbf{0.5962} \\ \hline
\multirow{3}{*}{DBLP\_v1}       & AUROC    & 0.7866 & \textbf{0.7973} & 0.6218          & \textbf{0.6825} & 0.6119          & \textbf{0.6885} & 0.6231 & \textbf{0.8044} & 0.7922          & \textbf{0.8006} & 0.8089          & \textbf{0.8222} \\
                                & AUPRC    & 0.8462 & \textbf{0.8515} & 0.7133          & \textbf{0.7769} & 0.7507          & \textbf{0.7796} & 0.7201 & \textbf{0.8626} & 0.8485          & \textbf{0.8537} & 0.8671          & \textbf{0.8716} \\
                                & F1-score & 0.7854 & \textbf{0.7974} & 0.6161          & \textbf{0.6805} & 0.5782          & \textbf{0.6868} & 0.5996 & \textbf{0.8028} & 0.7919          & \textbf{0.8007} & 0.8071          & \textbf{0.8220} \\  \hline \hline
\end{tabular}}
\vspace{-5mm}
\end{table}

\begin{table}[t]
\caption{Average AUROC, AUPRC, and F1-score on 12 datasets with multiple runs, using GAD models as baselines, where the white columns represent vanilla models and the gray ones represent models augmented by FracAug.}
\small
\centering
\scalebox{0.97}{
\setlength\tabcolsep{1pt}
\label{tab:gadperformance}
\begin{tabular}{cc|c
>{\columncolor[HTML]{DCDCDC}}c| c
>{\columncolor[HTML]{DCDCDC}}c |c
>{\columncolor[HTML]{DCDCDC}}c |c
>{\columncolor[HTML]{DCDCDC}}c } \hline \hline
Datasets                         & Metrics  & iGAD            & \multicolumn{1}{l|}{\cellcolor[HTML]{ffffff}{\scriptsize +FA}} & GmapAD          & \multicolumn{1}{l|}{\cellcolor[HTML]{ffffff}{\scriptsize +FA}} & RQGNN  & \multicolumn{1}{l|}{\cellcolor[HTML]{ffffff}{\scriptsize +FA}} & UniGAD          & \multicolumn{1}{l}{\cellcolor[HTML]{ffffff}{\scriptsize +FA}} \\ \hline
                                 \multirow{3}{*}{MCF-7}          & AUROC    & 0.5670          & \textbf{0.5756} & 0.5159          & \textbf{0.5342} & 0.5522 & \textbf{0.5709} & 0.5380          & \textbf{0.5480} \\
                                & AUPRC    & 0.3402          & \textbf{0.3525} & 0.3521          & \textbf{0.3577} & 0.2379 & \textbf{0.2648} & 0.2405          & \textbf{0.2527} \\
                                & F1-score & 0.4546          & \textbf{0.4549} & 0.3776          & \textbf{0.3961} & 0.5609 & \textbf{0.5804} & 0.4959          & \textbf{0.5008} \\ \hline
\multirow{3}{*}{MOLT-4}         & AUROC    & 0.5562          & \textbf{0.5573} & 0.5100          & \textbf{0.5374} & 0.5576 & \textbf{0.5696} & 0.5353          & \textbf{0.5445} \\
                                & AUPRC    & 0.3088          & \textbf{0.3092} & 0.2704          & \textbf{0.2827} & 0.2302 & \textbf{0.2468} & 0.2111          & \textbf{0.2241} \\
                                & F1-score & 0.4621          & \textbf{0.4636} & 0.4354          & \textbf{0.4592} & 0.5644 & \textbf{0.5718} & 0.5079          & \textbf{0.5125} \\ \hline
\multirow{3}{*}{PC-3}           & AUROC    & 0.5526          & \textbf{0.5674} & 0.5112          & \textbf{0.5266} & 0.5618 & \textbf{0.6043} & 0.5496          & \textbf{0.5559} \\
                                & AUPRC    & 0.1887          & \textbf{0.2159} & 0.2999          & \textbf{0.3094} & 0.2370 & \textbf{0.2663} & 0.3875          & \textbf{0.3889} \\
                                & F1-score & 0.5217          & \textbf{0.5229} & 0.3845          & \textbf{0.3938} & 0.5721 & \textbf{0.5972} & 0.3461          & \textbf{0.3542} \\ \hline
\multirow{3}{*}{SW-620}         & AUROC    & 0.5641          & \textbf{0.5776} & 0.5279          & \textbf{0.5362} & 0.5428 & \textbf{0.5692} & 0.5427          & \textbf{0.5688} \\
                                & AUPRC    & 0.3332          & \textbf{0.3701} & \textbf{0.3506} & 0.3479          & 0.1936 & \textbf{0.2219} & 0.2527          & \textbf{0.2585} \\
                                & F1-score & \textbf{0.4207} & 0.4029          & 0.3605          & \textbf{0.3734} & 0.5530 & \textbf{0.5654} & 0.4600          & \textbf{0.4903} \\ \hline
\multirow{3}{*}{NCI-H23}        & AUROC    & 0.5689          & \textbf{0.5721} & 0.5289          & \textbf{0.5489} & 0.5704 & \textbf{0.6061} & 0.5694          & \textbf{0.5860} \\
                                & AUPRC    & 0.2267          & \textbf{0.2288} & 0.3274          & \textbf{0.3401} & 0.2166 & \textbf{0.2500} & 0.2733          & \textbf{0.2756} \\
                                & F1-score & 0.5039          & \textbf{0.5066} & 0.3710          & \textbf{0.3827} & 0.5770 & \textbf{0.5820} & 0.4645          & \textbf{0.4831} \\ \hline
\multirow{3}{*}{OVCAR-8}        & AUROC    & 0.5609          & \textbf{0.5685} & 0.5209          & \textbf{0.5243} & 0.5549 & \textbf{0.5773} & 0.5360          & \textbf{0.5445} \\
                                & AUPRC    & 0.2319          & \textbf{0.2325} & \textbf{0.2863} & 0.2836          & 0.1933 & \textbf{0.2216} & \textbf{0.3196} & 0.3112          \\
                                & F1-score & 0.4880          & \textbf{0.4991} & 0.3992          & \textbf{0.4054} & 0.5618 & \textbf{0.5794} & 0.3873          & \textbf{0.4043} \\ \hline
\multirow{3}{*}{P388}           & AUROC    & 0.5143          & \textbf{0.5300} & 0.4782          & \textbf{0.5057} & 0.5952 & \textbf{0.6108} & 0.5104          & \textbf{0.5167} \\
                                & AUPRC    & 0.1923          & \textbf{0.1939} & 0.2478          & \textbf{0.2599} & 0.2484 & \textbf{0.2650} & 0.1679          & \textbf{0.1748} \\
                                & F1-score & 0.4669          & \textbf{0.4843} & 0.3894          & \textbf{0.4099} & 0.5879 & \textbf{0.5883} & 0.4781          & \textbf{0.4812} \\ \hline
\multirow{3}{*}{SF-295}         & AUROC    & 0.5811          & \textbf{0.5815} & 0.5414          & \textbf{0.5535} & 0.5582 & \textbf{0.5902} & 0.5439          & \textbf{0.5730} \\
                                & AUPRC    & 0.2666          & \textbf{0.2821} & 0.3066          & \textbf{0.3070} & 0.2141 & \textbf{0.2342} & \textbf{0.3000} & 0.2846          \\
                                & F1-score & \textbf{0.4836} & 0.4705          & 0.4030          & \textbf{0.4167} & 0.5719 & \textbf{0.5847} & 0.4117          & \textbf{0.4582} \\ \hline
\multirow{3}{*}{SN12C}          & AUROC    & 0.5522          & \textbf{0.5537} & 0.5343          & \textbf{0.5441} & 0.5597 & \textbf{0.6038} & 0.5433          & \textbf{0.5497} \\
                                & AUPRC    & 0.1817          & \textbf{0.1858} & 0.3262          & \textbf{0.3315} & 0.1927 & \textbf{0.2442} & \textbf{0.2028} & 0.1961          \\
                                & F1-score & \textbf{0.5151} & 0.5144          & 0.3754          & \textbf{0.3818} & 0.5648 & \textbf{0.5826} & 0.4851          & \textbf{0.4986} \\ \hline
\multirow{3}{*}{UACC257}        & AUROC    & 0.5697          & \textbf{0.5748} & 0.5394          & \textbf{0.5597} & 0.5528 & \textbf{0.5692} & 0.5710          & \textbf{0.5832} \\
                                & AUPRC    & 0.1906          & \textbf{0.1970} & 0.2927          & \textbf{0.3004} & 0.1601 & \textbf{0.1885} & 0.2510          & \textbf{0.2642} \\
                                & F1-score & 0.5201          & \textbf{0.5224} & 0.3986          & \textbf{0.4144} & 0.5522 & \textbf{0.5678} & 0.4676          & \textbf{0.4710} \\ \hline
\multirow{3}{*}{PROTEINS\_full} & AUROC    & 0.5976          & \textbf{0.6206} & 0.5041          & \textbf{0.6289} & 0.5641 & \textbf{0.6365} & 0.6173          & \textbf{0.6212} \\
                                & AUPRC    & 0.6200          & \textbf{0.6333} & 0.5169          & \textbf{0.6436} & 0.5673 & \textbf{0.6563} & 0.6295          & \textbf{0.6338} \\
                                & F1-score & 0.5960          & \textbf{0.6211} & 0.5020          & \textbf{0.6299} & 0.5600 & \textbf{0.6310} & 0.6178          & \textbf{0.6223} \\ \hline
\multirow{3}{*}{DBLP\_v1}       & AUROC    & 0.7755          & \textbf{0.7909} & 0.4975          & \textbf{0.5045} & 0.8065 & \textbf{0.8082} & 0.7601          & \textbf{0.7965} \\
                                & AUPRC    & 0.8377          & \textbf{0.8473} & 0.6242          & \textbf{0.6548} & 0.8584 & \textbf{0.8598} & 0.8346          & \textbf{0.8509} \\
                                & F1-score & 0.7749          & \textbf{0.7910} & 0.4968          & \textbf{0.5021} & 0.8060 & \textbf{0.8079} & 0.7549          & \textbf{0.7966} \\ \hline \hline
\end{tabular}}
\vspace{-4mm}
\end{table}
\begin{table}[t]
\caption{Average AUROC, AUPRC, and F1-score on 12 datasets with multiple runs, using graph-level augmentation models as baselines, where the white columns represent vanilla models and their own augmentation method, while the gray ones represent vanilla models augmented by FracAug. }
\small
\centering
\scalebox{0.755}{
\setlength\tabcolsep{1pt}
\label{tab:augperformance}
\begin{tabular}{cc|ccc
>{\columncolor[HTML]{DCDCDC}}c |cc
>{\columncolor[HTML]{DCDCDC}}c |cc
>{\columncolor[HTML]{DCDCDC}}c |cc
>{\columncolor[HTML]{DCDCDC}}c } \hline \hline
Datasets                         & Metrics  & MAAv            & NodeSam         & SubMix & \multicolumn{1}{l|}{\cellcolor[HTML]{ffffff}{\scriptsize +FA}} & GLAv            & GLA    & \multicolumn{1}{l|}{\cellcolor[HTML]{ffffff}{\scriptsize +FA}} & GMixupv & GMixup          & \multicolumn{1}{l|}{\cellcolor[HTML]{ffffff}{\scriptsize +FA}} & FGWMixupv & FGWMixup & \multicolumn{1}{l}{\cellcolor[HTML]{ffffff}{\scriptsize +FA}} \\ \hline
                                 \multirow{3}{*}{MCF-7}          & AUROC    & 0.5496 & \textbf{0.5727} & 0.5428          & 0.5695          & 0.5797 & 0.5735          & \textbf{0.6068} & 0.5730  & 0.5581          & \textbf{0.5935} & 0.5731          & 0.5502          & \textbf{0.5828} \\
                                & AUPRC    & 0.2346 & 0.2445          & 0.2224          & \textbf{0.2455} & 0.2558 & 0.2456          & \textbf{0.2944} & 0.2767  & 0.2864          & \textbf{0.3081} & 0.2720          & 0.2130          & \textbf{0.2945} \\
                                & F1-score & 0.5179 & 0.5635          & 0.5512          & \textbf{0.5721} & 0.5607 & 0.5609          & \textbf{0.5793} & 0.5186  & 0.4879          & \textbf{0.5220} & 0.5585          & 0.5283          & \textbf{0.5971} \\ \hline
\multirow{3}{*}{MOLT-4}         & AUROC    & 0.5506 & 0.5391          & 0.5113          & \textbf{0.5663} & 0.5585 & 0.5578          & \textbf{0.5792} & 0.5637  & 0.5547          & \textbf{0.5771} & 0.5477          & 0.5308          & \textbf{0.6067} \\
                                & AUPRC    & 0.2203 & 0.1914          & 0.1796          & \textbf{0.2356} & 0.2177 & 0.2159          & \textbf{0.2511} & 0.2411  & 0.2176          & \textbf{0.2559} & 0.1994          & 0.1939          & \textbf{0.3104} \\
                                & F1-score & 0.5595 & 0.5392          & 0.5048          & \textbf{0.5658} & 0.5540 & 0.5519          & \textbf{0.5733} & 0.5306  & 0.5368          & \textbf{0.5416} & \textbf{0.5446} & 0.5115          & 0.5366          \\ \hline
\multirow{3}{*}{PC-3}           & AUROC    & 0.5688 & 0.5805          & 0.5284          & \textbf{0.5821} & 0.6207 & 0.5938          & \textbf{0.6221} & 0.5705  & 0.5646          & \textbf{0.5782} & 0.5490          & 0.5442          & \textbf{0.6107} \\
                                & AUPRC    & 0.2112 & 0.2233          & 0.1854          & \textbf{0.2385} & 0.2770 & 0.2386          & \textbf{0.2790} & 0.3506  & 0.3456          & \textbf{0.3587} & 0.2028          & 0.2108          & \textbf{0.2623} \\
                                & F1-score & 0.5698 & 0.5685          & 0.5321          & \textbf{0.5848} & 0.5650 & 0.5569          & \textbf{0.5686} & 0.4085  & 0.4058          & \textbf{0.4101} & 0.5027          & 0.4914          & \textbf{0.5633} \\ \hline
\multirow{3}{*}{SW-620}         & AUROC    & 0.5577 & 0.5776          & 0.5232          & \textbf{0.5834} & 0.5822 & 0.5799          & \textbf{0.5936} & 0.5839  & 0.5722          & \textbf{0.5987} & 0.5265          & 0.5395          & \textbf{0.6183} \\
                                & AUPRC    & 0.2026 & 0.2205          & 0.1958          & \textbf{0.2279} & 0.2315 & 0.2258          & \textbf{0.2455} & 0.2599  & 0.2511          & \textbf{0.2728} & 0.1340          & 0.1777          & \textbf{0.2759} \\
                                & F1-score & 0.5641 & 0.5477          & 0.5273          & \textbf{0.5676} & 0.5422 & 0.5473          & \textbf{0.5786} & 0.5111  & 0.5015          & \textbf{0.5220} & 0.5283          & 0.5113          & \textbf{0.5729} \\ \hline
\multirow{3}{*}{NCI-H23}        & AUROC    & 0.5792 & 0.5634          & 0.5323          & \textbf{0.5979} & 0.5773 & 0.5762          & \textbf{0.6194} & 0.5912  & 0.5647          & \textbf{0.6014} & 0.5658          & 0.5760          & \textbf{0.6422} \\
                                & AUPRC    & 0.2273 & 0.1983          & 0.2017          & \textbf{0.2407} & 0.2082 & 0.2188          & \textbf{0.2680} & 0.2587  & 0.2139          & \textbf{0.2672} & 0.1986          & 0.2345          & \textbf{0.3105} \\
                                & F1-score & 0.5821 & 0.5625          & 0.5445          & \textbf{0.5835} & 0.5659 & 0.5779          & \textbf{0.5914} & 0.5058  & 0.5087          & \textbf{0.5134} & \textbf{0.5561} & 0.5066          & 0.5358          \\ \hline
\multirow{3}{*}{OVCAR-8}        & AUROC    & 0.5507 & 0.5494          & 0.5278          & \textbf{0.5726} & 0.5911 & 0.5859          & \textbf{0.6049} & 0.5786  & 0.5725          & \textbf{0.6024} & 0.5696          & 0.5713          & \textbf{0.6317} \\
                                & AUPRC    & 0.1775 & 0.1633          & 0.1665          & \textbf{0.2132} & 0.2346 & 0.2196          & \textbf{0.2447} & 0.2764  & 0.2994          & \textbf{0.3072} & 0.2696          & 0.2401          & \textbf{0.3060} \\
                                & F1-score & 0.5461 & 0.5408          & 0.5337          & \textbf{0.5749} & 0.5350 & 0.5548          & \textbf{0.5728} & 0.4733  & 0.4469          & \textbf{0.4766} & 0.4831          & 0.4950          & \textbf{0.5211} \\ \hline
\multirow{3}{*}{P388}           & AUROC    & 0.5500 & 0.5069          & 0.5057          & \textbf{0.5720} & 0.5622 & 0.5816          & \textbf{0.6057} & 0.5469  & 0.5265          & \textbf{0.5647} & 0.5480          & 0.5409          & \textbf{0.5729} \\
                                & AUPRC    & 0.1958 & 0.1985          & 0.1127          & \textbf{0.2229} & 0.2157 & 0.2264          & \textbf{0.2632} & 0.1694  & 0.1536          & \textbf{0.1957} & 0.2056          & 0.1951          & \textbf{0.2362} \\
                                & F1-score & 0.5520 & 0.5000          & 0.4987          & \textbf{0.5746} & 0.5372 & 0.5766          & \textbf{0.5925} & 0.5315  & 0.5078          & \textbf{0.5373} & 0.5421          & 0.5282          & \textbf{0.5798} \\ \hline
\multirow{3}{*}{SF-295}         & AUROC    & 0.5649 & 0.5579          & 0.5292          & \textbf{0.5753} & 0.5954 & 0.6060          & \textbf{0.6197} & 0.5665  & 0.5687          & \textbf{0.6040} & 0.5893          & 0.5981          & \textbf{0.6459} \\
                                & AUPRC    & 0.2114 & 0.1939          & 0.2218          & \textbf{0.2252} & 0.2316 & 0.2451          & \textbf{0.2648} & 0.2004  & 0.2061          & \textbf{0.2509} & 0.2331          & 0.2585          & \textbf{0.3017} \\
                                & F1-score & 0.5736 & 0.5644          & 0.5406          & \textbf{0.5820} & 0.5643 & 0.5702          & \textbf{0.5855} & 0.5245  & 0.5205          & \textbf{0.5371} & 0.5300          & 0.5161          & \textbf{0.5623} \\ \hline
\multirow{3}{*}{SN12C}          & AUROC    & 0.5509 & 0.5639          & 0.5160          & \textbf{0.5795} & 0.5715 & 0.6003          & \textbf{0.6141} & 0.5713  & 0.5336          & \textbf{0.5984} & 0.5831          & 0.5693          & \textbf{0.6314} \\
                                & AUPRC    & 0.1726 & 0.1845          & 0.1966          & \textbf{0.2164} & 0.2048 & 0.2344          & \textbf{0.2528} & 0.2163  & 0.2047          & \textbf{0.2524} & 0.2382          & 0.2252          & \textbf{0.2929} \\
                                & F1-score & 0.5538 & 0.5405          & 0.5190          & \textbf{0.5770} & 0.5644 & 0.5590          & \textbf{0.5655} & 0.5136  & 0.4920          & \textbf{0.5195} & 0.5085          & 0.5002          & \textbf{0.5326} \\ \hline
\multirow{3}{*}{UACC257}        & AUROC    & 0.5623 & 0.5023          & 0.5211          & \textbf{0.5947} & 0.6159 & 0.6198          & \textbf{0.6327} & 0.5853  & 0.5843          & \textbf{0.6209} & 0.5535          & 0.5632          & \textbf{0.6368} \\
                                & AUPRC    & 0.1805 & 0.0559          & 0.0942          & \textbf{0.2210} & 0.2755 & 0.2745          & \textbf{0.2978} & 0.2365  & 0.2285          & \textbf{0.2963} & 0.1534          & 0.2172          & \textbf{0.2718} \\
                                & F1-score & 0.5541 & 0.5005          & 0.5214          & \textbf{0.5784} & 0.5033 & \textbf{0.5131} & 0.5050          & 0.4993  & \textbf{0.5042} & 0.4898          & 0.5470          & 0.4830          & \textbf{0.5571} \\ \hline
\multirow{3}{*}{PROTEINS\_full} & AUROC    & 0.6009 & 0.6083          & 0.4998          & \textbf{0.6217} & 0.5652 & 0.5325          & \textbf{0.6249} & 0.5411  & 0.5132          & \textbf{0.6097} & 0.5078          & 0.5259          & \textbf{0.6082} \\
                                & AUPRC    & 0.6183 & 0.6294          & 0.6186          & \textbf{0.6366} & 0.6053 & 0.5343          & \textbf{0.6476} & 0.5810  & 0.5057          & \textbf{0.6244} & 0.5300          & \textbf{0.6934} & 0.6333          \\
                                & F1-score & 0.6015 & 0.6039          & 0.4316          & \textbf{0.6214} & 0.5577 & 0.5291          & \textbf{0.6227} & 0.5348  & 0.5025          & \textbf{0.6102} & 0.4909          & 0.3809          & \textbf{0.6031} \\ \hline
\multirow{3}{*}{DBLP\_v1}       & AUROC    & 0.6446 & 0.6608          & 0.6205          & \textbf{0.6822} & 0.7040 & 0.6402          & \textbf{0.7222} & 0.7939  & 0.7885          & \textbf{0.7994} & 0.7865          & 0.7772          & \textbf{0.7989} \\
                                & AUPRC    & 0.7689 & 0.7868          & \textbf{0.7947} & 0.7816          & 0.7882 & 0.7450          & \textbf{0.8085} & 0.8503  & 0.8471          & \textbf{0.8563} & 0.8461          & 0.8377          & \textbf{0.8549} \\
                                & F1-score & 0.6252 & 0.6408          & 0.6291          & \textbf{0.6778} & 0.7029 & 0.6147          & \textbf{0.7177} & 0.7937  & 0.7878          & \textbf{0.7985} & 0.7856          & 0.7772          & \textbf{0.7981} \\ \hline \hline
\end{tabular}}
\vspace{-5mm}
\end{table}

\subsection{Experimental Setup}
\label{subsec:experimentalsetup}
\textbf{Datasets.} We evaluate FracAug on 12 real-world datasets, including MCF-7, MOLT-4, PC-3, SW-620, NCI-H23, OVCAR-8, P388, SF-295, SN12C, UACC257, PROTEINS\_full and DBLP\_v1.  These datasets are obtained from TUDataset\footnote{https://chrsmrrs.github.io/datasets/docs/datasets/}, and their detailed statistics are listed in Appendix \ref{app:datasetandbaseline}. We randomly divide each dataset into 1\%/1\%/98\% for $\mathcal{D}_{train}\text{/}\mathcal{D}_{val}\text{/}\mathcal{D}_{test}$ to simulate the limited supervision scenario in real applications.

\textbf{Baselines.} We integrate our FracAug with 10 distinct GNNs, including generalized graph classification models and specialized GAD models, to demonstrate its broad applicability. Besides, to further confirm the usefulness of FracAug, we compare FracAug against 4 SOTA graph-level augmentation frameworks based on their original vanilla models. 
\begin{itemize} [topsep=0.5mm, partopsep=0pt, itemsep=0pt, leftmargin=10pt]
    \item Graph Classification: GCN \citep{gcn17kipf}, GraphSAGE \citep{graphsage17hamilton}, GAT \citep{gat18velickovic}, GIN \citep{gin19xu}, LRGNN \citep{lrgnn23wei}, and GRDL \citep{grdl24wang}. 
    \item Graph-level Anomaly Detection: iGAD \citep{igad22zhang}, GmapAD \citep{gmapad23ma}, RQGNN \citep{rqgnn24dong}, and UniGAD \citep{unigad24lin}. 
    \item Graph-level Augmentation: MAA \citep{maa22yoo}, GLA \citep{gla22yue}, GMixup \citep{gmixup22han}, and FGWMixup \citep{fgwmixup23ma}. 
\end{itemize}

\textbf{Experimental Settings.} To ensure fair evaluation, we standardize evaluations by: (1) sourcing all baseline code from GitHub and replacing loss functions with weighted version to mitigate class imbalance; (2) using authors’ recommended hyperparameters for baselines, while optimizing FracAug’s hyperparameters via grid search to maximize the summed AUROC/AUPRC/F1-score on validation sets. Complete configurations are detailed in Appendix \ref{app:experimentsetting}.

\subsection{Experimental Results}
\label{subsec:experimentalresults}

We evaluate the performance of FracAug on 6 graph classification models and 4 GAD models. Tables \ref{tab:gcperformance} and \ref{tab:gadperformance} report the AUROC, AUPRC, and F1-score on 12 datasets. Besides, we also compare FracAug with 4 graph-level augmentation methods on their vanilla models, as shown in Table \ref{tab:augperformance}. The best performance of each model is highlighted in boldface. To sum up, FracAug effectively boosts the performance of GNNs and outperforms almost all baselines on these real-world datasets. Next, we provide our detailed observations. 

{\bf Augmentation for Graph Classification Models.} We analyze 4 generalized GNNs (GCN, GraphSAGE, GAT, and GIN) and 2 recent models (LRGNN and GRDL) under limited supervision conditions. While generalized GNNs, due to architectural simplicity, struggle to capture nuanced anomaly patterns in GAD tasks, FracAug boosts their performance across most datasets as shown in Table \ref{tab:gcperformance}, validating its augmentation efficacy. Surprisingly, LRGNN and GRDL initially underperform simpler GNNs in some cases, likely hindered by label scarcity, but regain competitiveness when integrated with FracAug, highlighting FracAug's adaptability to advanced architectures. 

{\bf Augmentation for Graph-level Anomaly Detection Models.} Specialized GAD models (iGAD, GmapAD, RQGNN, and UniGAD) exploit task-specific properties but falter under limited supervision due to insufficient generalization capability. Notably, these task-specific architectures may underperform even basic GNNs in low-label regimes as presented in Table \ref{tab:gadperformance}, emphasizing FracAug’s effectiveness. Our framework universally elevates their performance by compensating for supervision scarcity, validating its versatility across model paradigms.

{\bf Comparison with Graph-level Augmentation Frameworks.} To further prove the effectiveness of FracAug, we compare it against leading graph-level augmentation frameworks, including MAA, GLA, GMixup, and FGWMixup. In Table \ref{tab:augperformance}, we denote their corresponding vanilla models as MAAv, GLAv, GMixupv, and FGWMixupv, respectively. Such a setting will preserve the ability of those augmentation frameworks. Nevertheless, as we can see, the augmentation methods fail to generalize effectively to GAD tasks under limited supervision---the performance of the vanilla models may drop after the augmentation. In contrast, our FracAug can boost all vanilla models across real-world datasets, which demonstrates the usefulness of FracAug. 

Beyond these primary results, we provide extensive additional analyses: complexity analysis, hyperparameter analysis, ablation study, loss function comparison, performance with more training data, and the learned parameters on different datasets in Appendix \ref{app:complexityanalysis}, \ref{app:hyperparameteranalysis}, \ref{app:ablationstudy}, \ref{app:lossfunctioncomoparson}, \ref{app:performancewithmoredata}, and \ref{app:learnedparameter}, separately. These supplementary analyses can further demonstrate the effectiveness of our proposed FracAug. 

\section{Conclusion}
\label{sec:conclusion}
In this paper, we investigate the efficacy of leveraging fractional graph variants for data augmentation in GAD under limited supervision scenarios. Based on the analysis, we design a model-agnostic plug-in augmentation framework, FracAug, which includes three key components: FGG, WDML, and MVP. FGG with WDML captures semantics from original samples and then generates semantic-preserving fractional graphs during model training, unaffected by the imbalanced data distribution, while MVP employs mutual verification to enhance pseudo-labeling reliability, iteratively expanding the training set. Comprehensive experiments demonstrate that FracAug not only effectively improves the performance of any given GNN but also significantly outperforms other graph-level augmentation methods, demonstrating the effectiveness of our method.

\newpage
\bibliography{iclr2026_conference}
\bibliographystyle{iclr2026_conference}

\newpage

\appendix

\section{Proofs}
\label{app:proof}
{\bf Proof of Theorem 1.}
To derive the approximation of $\vect{A}^\alpha$ and the corresponding error bound, we first consider a function for real numbers, i.e., $f(x)=x^\alpha$ defined on an interval $[a, b]\subset (0, +\infty)$. To satisfy the requirement of Chebyshev series approximation, we map $[a, b]$ to the standard Chebyshev interval $[-1, 1]$ via the linear transformation: 
\begin{equation*}
    x = \frac{2}{b-a}(x'-\frac{b+a}{2}),
\end{equation*}
where $x'\in[a, b]$ maps to $x\in [-1, 1]$. Then we further define: 
\begin{equation*}
    \tilde{f}(x)=(\frac{(b-a)x+(b+a)}{2})^\alpha, 
\end{equation*}
which can be approximated using Chebyshev series approximation as: 
\begin{equation*}
    \tilde{f}(x)\approx p_T(x)=\sum_{t=0}^Tc_tP_t(x), 
\end{equation*}
where $P_t(x)$ is the $t$-th Chebyshev polynomial, and the $c_t$ are the Chebyshev coefficients. Specifically, we can find the coefficients $c_t$ through the application of an inner product: 
\begin{equation*}
    \int_{-1}^{+1}\frac{P_m(x)\tilde{f}(x)}{\sqrt{1-x^2}}dx=\sum_{t=0}^\infty c_t\int_{-1}^{+1}\frac{P_m(x)P_t(x)}{\sqrt{1-x^2}}dx. 
\end{equation*}
On the interval $[-1, 1]$, we have: 
\begin{equation*}
    \int_{-1}^{+1}\frac{P_m(x)P_t(x)}{\sqrt{1-x^2}}dx = \left\{ \begin{aligned}
        &0,\ \ \ m\neq t, \\
        &\pi,\ \ \ m=t=0, \\
        &\frac{\pi}{2}, \ \ \ m=t\neq 0, 
    \end{aligned}
    \right. 
\end{equation*}
so we can derive: 
\begin{equation*}
    c_t=\left\{ \begin{aligned}
        \frac{1}{\pi}\int_{-1}^{+1}\frac{P_t(x)\tilde{f}(x)}{\sqrt{1-x^2}}dx,\ \ \ t=0, \\
        \frac{2}{\pi}\int_{-1}^{+1}\frac{P_t(x)\tilde{f}(x)}{\sqrt{1-x^2}}dx, \ \ \ t\neq 0. 
    \end{aligned}
    \right. 
\end{equation*}

Afterward, to obtain the error bound of the approximation, we leverage the following Theorem: 
\begin{theorem} \label{thm:convergeanalytic}
    (Theorems 8.1 and 8.2 from previous work \citep{convergeanalytic19trefethen}) Let a function $f(x)$ analytic in $[-1, 1]$ be analytically continuable to open Bernstein ellipse $\vect{E}_p$, where it satisfies $|f(x)|\leq M$ for some $M$, then for each $t\geq 0$, its Chebyshev approximation $p_T(x)$ satisfies $||f(x)-p_T(x)||\leq \frac{4M\rho^{-T}}{\rho-1}$, where $\rho$ depends on the distance from $[-1, 1]$ to the nearest singularity of $f(x)$. 
\end{theorem}

The function $f(x)=x^\alpha$ has a branch point at $x=0$. For $[a,b]\subset (0, +\infty)$, the mapped function $\tilde{f}(x)$ is analytic in a Bernstein ellipse $\vect{E}_p$, excluding $x=0$. Therefore, $\tilde{f}(x)$ satisfies Theorem \ref{thm:convergeanalytic}, so we can have: 
\begin{equation*}
    ||\tilde{f}(x)-p_T(x)||\leq \frac{4M\rho^{-T}}{\rho-1}=\beta e^{-\gamma T}, 
\end{equation*}
where $\beta=\frac{4M}{\rho-1}$ and $\gamma=\ln\rho$. 

Similarly, we can directly apply the function to $\vect{A}$ with eigenvalues in $[\lambda_{min}, \lambda_{max}]\subset (0, +\infty)$, then we can conclude: 
\begin{equation*}
    ||\vect{A}^\alpha-p_T(\vect{A})||\leq \beta e^{-\gamma T}, 
\end{equation*}
where $\beta, \gamma$ is derived from $[\lambda_{min}, \lambda_{max}]$. {\hfill \qedsymbol}

{\bf Proof of Theorem 2.} Using the Dunford-Taylor integral, for a contour $\Gamma$ enclosing the spectra of $\vect{A}$ and $\vect{A}+\vect{P}$, we have: 
\begin{equation*}
    \begin{aligned}
        \vect{A}^\alpha&=\frac{1}{2\pi i}\int_{\Gamma}x^\alpha(x\vect{I}-\vect{A})^{-1}dx,\\
        (\vect{A}+\vect{P})^\alpha&=\frac{1}{2\pi i}\int_{\Gamma}x^\alpha(x\vect{I}-(\vect{A}+\vect{P}))^{-1}dx. 
    \end{aligned}
\end{equation*}

Then we subtract the two integrals: 
\begin{equation*}
    \vect{A}^\alpha-(\vect{A}+\vect{P})^\alpha=\frac{1}{2\pi i}\int_{\Gamma}x^\alpha[(x\vect{I}-\vect{A})^{-1}-(x\vect{I}-(\vect{A}+\vect{P}))^{-1}]dx. 
\end{equation*}

After applying the resolvent identity, we can have: 
\begin{equation*}
    (x\vect{I}-\vect{A})^{-1}-(x\vect{I}-(\vect{A}+\vect{P}))^{-1}=(x\vect{I}-\vect{A})^{-1}\vect{P}(x\vect{I}-(\vect{A}+\vect{P}))^{-1}. 
\end{equation*}

By substituting back into the integral, we have: 
\begin{equation*}
    \vect{A}^\alpha-(\vect{A}+\vect{P})^\alpha=\frac{1}{2\pi i}\int_{\Gamma}x^\alpha(x\vect{I}-\vect{A})^{-1}\vect{P}(x\vect{I}-(\vect{A}+\vect{P}))^{-1}dx.
\end{equation*}

Take the operator norm and apply submultiplicativity: 
\begin{equation*}
    ||\vect{A}^\alpha-(\vect{A}+\vect{P})^\alpha||\leq\frac{1}{2\pi}\int_{\Gamma}|x^\alpha|||(x\vect{I}-\vect{A})^{-1}||||\vect{P}||||(x\vect{I}-(\vect{A}+\vect{P}))^{-1}|||dx|.
\end{equation*}

If we choose $\Gamma$ to be a contour at distance $d>0$ from the spectra of $\vect{A}$, we can have: 
\begin{equation*}
    (x\vect{I}-\vect{A})^{-1}=\vect{U}(x\vect{I}-\vect{\Lambda})^{-1}\vect{U}^T, 
\end{equation*}
where $\vect{A}=\vect{U}\vect{\Lambda}\vect{U}^T$ is the eigendecomposition of $\vect{A}$ and the corresponding norm is: 
\begin{equation*}
    ||(x\vect{I}-\vect{A})^{-1}||=||(x\vect{I}-\vect{\Lambda})^{-1}||=\max_{\lambda\in\sigma(\vect{A})}\frac{1}{|x-\lambda|}=\frac{1}{\text{dist}(x, \sigma(\vect{A}))}=\frac{1}{d}, 
\end{equation*}
where $\sigma(\vect{A})$ is the spectrum of $\vect{A}$ and $\text{dist}(\cdot)$ is the distance function. 

For a small perturbation $||\vect{P}||$, the spectrum of $\vect{A}+\vect{P}$ will lie in a neighborhood of the spectrum of $\vect{A}$. Specifically, for any eigenvalue $\lambda'$ of $\vect{A}+\vect{P}$, there exists an eigenvalue $\lambda$ of $\vect{A}$ such that $|\lambda'-\lambda|\leq||\vect{P}||$, which implies: 
\begin{equation*}
    \text{dist}(x, \sigma(\vect{A}+\vect{P}))\geq \text{dist}(x, \sigma(\vect{A}))-||\vect{P}||. 
\end{equation*}

Then for $x\notin\sigma(\vect{A}+\vect{P})$, we use the Neumann series: 
\begin{equation*}
    (x\vect{I}-(\vect{A}+\vect{P}))^{-1}=(x\vect{I}-\vect{A})^{-1}\sum_{i=0}^{+\infty}[\vect{P}(x\vect{I}-\vect{A})^{-1}]^i, 
\end{equation*}
which converges if $||\vect{P}(x\vect{I}-\vect{A})^{-1}||<1$. 

Afterward, we take its norm and apply submultiplicativity: 
\begin{equation*}
    \begin{aligned}
        ||(x\vect{I}-(\vect{A}+\vect{P}))^{-1}||&\leq\frac{||(x\vect{I}-\vect{A})^{-1}||}{1-||\vect{P}||||(x\vect{I}-\vect{A})^{-1}||} \\
        &=\frac{1}{\text{dist}(x, \sigma(\vect{A}))-||\vect{P}||}\\
        &\leq\frac{1}{d}. 
    \end{aligned}
\end{equation*}

Then let $M=\max_{x\in\Gamma}|x^\alpha|$ and $\text{L}(\cdot)$ be the length function, we can have: 
\begin{equation*}
    ||\vect{A}^\alpha-(\vect{A}+\vect{P})^\alpha||\leq\frac{1}{2\pi d^2}M||\vect{P}||\text{L}(\Gamma)
\end{equation*}

Define $c=\frac{ML(\Gamma)}{2\pi d^2}$, yielding $||\vect{A}^\alpha-(\vect{A}+\vect{P})^\alpha||\leq c||\vect{P}||$. 

Besides, for a generated adjacency matrix from the perturbation method, it can be diagonalized, so we can have: 
\begin{equation*}
    (\vect{A}+\vect{P})-(\vect{A}+\vect{P})^\alpha=\vect{V}(\vect{\Sigma}-\vect{\Sigma}^\alpha)\vect{V}^T, 
\end{equation*}
where $\vect{A}+\vect{P}=\vect{V}\vect{\Sigma}\vect{V}^T$ and $\vect{\Sigma}$ is a diagonal matrix composed of $(\lambda_1, \lambda_2, \cdots, \lambda_n)$. Take the norm, we can get: 
\begin{equation*}
    ||(\vect{A}+\vect{P})-(\vect{A}+\vect{P})^\alpha||=\max_i|\lambda_i-\lambda_i^\alpha|. 
\end{equation*}

Finally, by applying the submultiplicativity, we can conclude: 
\begin{equation*}
    ||\vect{A}^\alpha - (\vect{A}+\vect{P})||\leq c||\vect{P}||+\max_i|\lambda_i-\lambda_i^\alpha|, 
\end{equation*}
where $c$ depends on $\alpha$ and the spectral gap of $\vect{P}$, and $\lambda_i$ is the $i$-th eigenvalue of $\vect{A} + \vect{P}$. {\hfill \qedsymbol}

{\bf Proof of Proposition 1.} Assume the two prediction error rates for original graphs and corresponding fractional graphs are two Bernoulli variables with mean $\delta$, and the correlation of the errors is $\rho$. Then we have the joint error rate: 
\begin{equation*}
    \mathbb{P}(\text{Both wrong})=\delta^2+\rho \delta(1-\delta). 
\end{equation*}
Since the original error rate is $\delta$, the mutual verification will lower the error with the reduction factor $\delta+\rho\delta(1-\delta)$. 

According to the above analysis, the error rate of mutual verification is $p=\delta^2+\rho \delta(1-\delta)$. Assuming it is also a Bernoulli variable, the variance can be calculated as: 
\begin{equation*}
    v=(\delta^2+\rho\delta(1-\delta))(1-\delta^2-\rho\delta(1-\delta)). 
\end{equation*}

For a small error rate $\delta$, we can approximate it as $v=\rho\delta(1-\rho\delta)$. Therefore, the reduction factor of variance is close to $\rho$. {\hfill \qedsymbol}

\section{Datasets and Baselines}
\label{app:datasetandbaseline}

\begin{table}[t]
\caption{Statistics of 12 real-world datasets, where $n_n$ is the number of normal graphs, $n_a$ is the number of anomalous graphs, $h=\frac{n_a}{n_n+n_a}$ is the anomalous ratio, $\bar{n}$ is the average number of nodes, $\bar{m}$ is the average number of edges, and $F$ is the number of attributes.}
\small
\centering
\scalebox{0.86}{
\setlength\tabcolsep{1.5pt}
\label{tab:datasets}
\begin{tabular}{ccccccccccccc}
\hline \hline
Dataset	&	MCF-7	&	MOLT-4	&	PC-3	&	SW-620	&	NCI-H23	&	OVCAR-8	&	P388	&	SF-295	&	SN12C	&	UACC257	& PROTEINS\_full & DBLP\_v1 \\
\midrule
$n_n$	&	25476	&	36625	&	25941	&	38122	&	38296	&	38437	&	39174	&	38246	&	38049	&	38345	& 663 & 9926 \\
$n_a$	&	2294	&	3140	&	1568	&	2410	&	2057	&	2079	&	2298	&	2025	&	1955	&	1643 	& 450 & 9530 \\
$h$	&	0.0826	&	0.079	&	0.057	&	0.0595	&	0.051	&	0.0513	&	0.0554	&	0.0503	&	0.0489	&	0.0411	& 0.4043 & 0.4898 \\
$\bar{n}$	&	26.4	&	26.1	&	26.36	&	26.06	&	26.07	&	26.08	&	22.11	&	26.06	&	26.08	&	262.09	& 39.06 & 10.48 \\
$\bar{m}$	&	28.53	&	28.14	&	28.49	&	28.09	&	28.1	&	28.11	&	23.56	&	28.09	&	28.11	&	28.13	& 72.82 & 19.65 \\
$F$	&	46	&	64	&	45	&	65	&	65	&	65	&	72	&	65	&	65	&	64	& 3 & 41325 \\
\hline \hline
\vspace{-4mm}
\end{tabular}
}
\end{table}

\textbf{Datasets.} The datasets used in our experiments are collected by TUDataset \citep{tudataset20morris}. Specifically, MCF-7, MOLT-4, PC-3, SW-620, NCI-H23, OVCAR-8, P388, SF-295, SN12C, and UACC257 are small-molecule datasets from PubChem \footnote{https://pubchem.ncbi.nlm.nih.gov/}, which provide information on the biological activities of small molecules. In these datasets, nodes represent atoms within chemical compounds, while edges indicate the chemical bonds connecting pairs of atoms. Each dataset corresponds to a specific type of cancer screening, with outcomes classified as either active or inactive. We consider inactive chemical compounds as normal graphs and active compounds as anomalous graphs. Furthermore, the attributes are derived from node labels using one-hot encoding. 

Besides, PROTEINS\_full is a typical bioinformatics-related dataset \citep{protein03dobson}, which processes several proteins represented as graphs. In this dataset, nodes and edges are formulated in a similar way to small-molecule datasets from PubChem. This dataset aims to classify enzymes and non-enzymes, which are denoted as normal and anomalous graphs, respectively. 

Beyond the above datasets, we also conduct experiments on DBLP\_v1 \citep{dblp13pan}, which consists of bibliography data in computer science. Each record in DBLP\_v1 is associated with a number of attributes such as abstract, authors, year, venue, title, and reference ID. Since the dimension of the attributes is high, we first utilize EVD to lower the dimension to 16. In this dataset, nodes denote papers, while edges represent reference relations between papers. The classification task is to predict whether a paper belongs to the CVPR (computer vision and pattern recognition) or DBDM (database and data mining) conferences, which are seen as normal and anomalous, respectively. 

\textbf{Baselines.}  The first group is graph classification models:
\begin{itemize}[topsep=0.5mm, partopsep=0pt, itemsep=0pt, leftmargin=10pt]
    \item GCN \citep{gcn17kipf}: a GNN that uses a convolution function on a graph to propagate information within the neighborhood of nodes;
    \item GraphSAGE \citep{graphsage17hamilton}: a GNN that leverages a sampling technique to aggregate features from the neighborhood. 
    \item GAT \citep{gat18velickovic}: a GNN that adopts an attention mechanism within the neighborhood of each node;
    \item GIN \citep{gin19xu}: a GNN that follows graph isomorphism to capture the properties of a graph. 
    \item LRGNN \citep{lrgnn23wei}: a GNN stacking multiple GNNs to extract the long-range dependencies;
    \item GRDL \citep{grdl24wang}: a GNN treating node embeddings as a discrete distribution, enabling direct classification without global pooling. 
\end{itemize}
The second group is GAD models:
\begin{itemize}[topsep=0.5mm, partopsep=0pt, itemsep=0pt, leftmargin=10pt]
    \item iGAD \citep{igad22zhang}: a GNN with a substructure-aware component to capture properties of anomalous graphs. 
    \item GmapAD \citep{gmapad23ma}: a GNN mapping graphs into a latent space where anomalies can be effectively detected;
    \item RQGNN \citep{rqgnn24dong}: a GNN using Rayleigh Quotient to obtain information from both spectral and spatial spaces.
    \item UniGAD \citep{unigad24lin}: a GNN that unifies different levels of graph-related tasks. 
\end{itemize}
The third group is graph-level augmentation frameworks:
\begin{itemize}[topsep=0.5mm, partopsep=0pt, itemsep=0pt, leftmargin=10pt]
    \item MAA \citep{maa22yoo}: a framework using node split and merge, and subgraph mix to augment graphs heuristically; 
    \item GLA \citep{gla22yue}: a framework augmenting data in the representation space from the most difficult direction while keeping the label of augmented data the same as the original samples; 
    \item GMixup \citep{gmixup22han}: a framework that interpolates graphons of different classes in the Euclidean space to get mixed graphons;
    \item FGWMixup \citep{fgwmixup23ma}: a framework that seeks a midpoint of source graphs in the Fused Gromov-Wasserstein metric space to interpolate graphons of different classes. 
\end{itemize}

\section{Algorithm}
\label{app:algorithm}

\IncMargin{1em}
\begin{algorithm}

\caption{Preprocess}\label{alg:preprocess}
\KwIn{$\mathcal{D}, k_l, k_s$}
\For{$G$ in $\mathcal{D}$}{
    $G.\vect{A}\leftarrow\frac{1}{2}(\vect{I}+G.\vect{D}^{-\frac{1}{2}}*G.\vect{A}*G.\vect{D}^{-\frac{1}{2}})$\;
    $G.\vect{U}_l, G.\vect{\Lambda}_l\leftarrow\text{EVD}(G.\vect{A}, k_l)$\;
    $G.\vect{U}_s, G.\vect{\Lambda}_s\leftarrow\text{EVD}(G.\vect{A}, k_s)$\;
}

\end{algorithm}
\DecMargin{1em}
\IncMargin{1em}
\begin{algorithm}

\caption{FGG}\label{alg:fgg}
\KwIn{$\mathcal{D}, H_l, H_s$}
\KwOut{$\mathcal{D}'$}
\For{$G$ in $\mathcal{D}$}{
    \For{$i=0$ to $H_l$}{
        $G_l\leftarrow G_l+\vect{\omega}_l[i]*G.\vect{U}_l*G.\vect{\Lambda}_l^{\vect{\alpha}_l[i]}*G.\vect{U}_l^T$\; 
    }
    \For{$i=0$ to $H_l$}{
        $G_s\leftarrow G_s+\vect{\omega}_s[i]*G.\vect{U}_s*G.\vect{\Lambda}_s^{\vect{\alpha}_s[i]}*G.\vect{U}_s^T$\; 
    }
    $G'\leftarrow\omega*G_l+(1-\omega)*G_s$\;
    $\mathcal{D}'\leftarrow\mathcal{D}'\cup G'$\;
}
Return $\mathcal{D}'$\;
\end{algorithm}
\DecMargin{1em}
\IncMargin{1em}
\begin{algorithm}

\caption{WDML}\label{alg:wdml}
\KwIn{$f, \mathcal{D}, \mathcal{D}'$}
\For{$G, G'$ in $\mathcal{D}, \mathcal{D}'$}{
    $\vect{s}, \vect{o}\leftarrow f(G)$\;
    $\vect{s}', \vect{o}'\leftarrow f(G')$\;
    $m\leftarrow\frac{1-\text{cos}(\vect{o}, \vect{o}')}{2}$\;
    $L_{\text{WDML}}\leftarrow L_{\text{WDML}}+-\frac{1}{N_{G.y}}\log\frac{e^{\vect{s}[G.y]-m}}{e^{\vect{s}[G.y]-m}+e^{\vect{s}[1-G.y]}}$\;
}
$L_{\text{WDML}}.\text{backward()}$\;
\end{algorithm}
\DecMargin{1em}
\IncMargin{1em}
\begin{algorithm}

\caption{MVP}\label{alg:mvp}
\KwIn{$f, \mathcal{D}, \mathcal{D}', \tau_n, \tau_a$}
\KwOut{$\mathcal{D}''$}
\For{$G, G'$ in $\mathcal{D}, \mathcal{D}'$}{
    $\vect{s}, \vect{o}\leftarrow f(G)$\;
    $\vect{s}', \vect{o}'\leftarrow f(G')$\;
    \If {$\vect{s}[0]<\tau_n\land \vect{s}'[0]\leq\tau_n$} {
    $G.y\leftarrow 0$\;
    $\mathcal{D}''\leftarrow\mathcal{D}''\cup G$\;
    }
    \ElseIf {$\vect{s}[1]<\tau_a\land \vect{s}'[1]\leq\tau_a$} {
        $G.y\leftarrow 1$\;
    $\mathcal{D}''\leftarrow\mathcal{D}''\cup G$\;
    }
}
Return $\mathcal{D}''$\;
\end{algorithm}
\DecMargin{1em}
\IncMargin{1em}
\begin{algorithm}

\caption{FracAug}\label{alg:fracaug}
\KwIn{$f, \mathcal{D}_{train}, \mathcal{D}_{val}, \mathcal{D}_{test}, H_l, H_s, k_l, k_s, e_{warmup}, e_{aug}, \tau_n, \tau_a$}

Preprocess($\mathcal{D}_{train}\cup\mathcal{D}_{val}\cup\mathcal{D}_{test}, k_l, k_s$)\;
$D_{train}'\leftarrow D_{train}$\;
\For{$e=0$ to $e_f$}{
     \If {$e>e_{warmup}\land e \% e_{aug} == 0$}{
        \For{$e'=0$ to $e_{\text{FGG}}$}{
            $\mathcal{D}_{temp}\leftarrow 
            \text{FGG}(\mathcal{D}_{train}, H_l, H_s)$\;
            $\text{WDML}(f, \mathcal{D}, \mathcal{D}_{temp})$\;
        }
        $\mathcal{D}_{temp}\leftarrow 
        \text{FGG}(\mathcal{D}_{val}\cup\mathcal{D}_{test}, H_l, H_s)$\;
        $\mathcal{D}_{temp}\leftarrow\text{MVP}(f, \mathcal{D}_{val}\cup\mathcal{D}_{test}, \mathcal{D}_{temp}, \tau_n, \tau_a)$\;
        $D_{train}'\leftarrow D_{train}\cup \mathcal{D}_{temp}$\;
     }
     train($f, D_{train}')$)\;
}
\end{algorithm}
\DecMargin{1em}

\section{Complexity Analysis}
\label{app:complexityanalysis}
\begin{table}[t]
\caption{Comparison of average running time.}
\small
\centering
\scalebox{0.82}{
\setlength\tabcolsep{1pt}
\label{tab:runningtime}
\begin{tabular}{c|ccc|cc|cc|cc} \hline \hline
Datasets       & NSv+FA        & NodeSam & SubMix & GLAv+FA       & GLA     & GMixupv+FA    & GMixup & FGWMixupv+FA  & FGWMixup \\  \hline
MCF-7          & 92.18+58.56   & 1282.84 & 876.39 & 92.00+73.57   & 1493.55 & 90.54+107.51  & 51.64  & 92.29+106.76  & 553.90   \\
MOLT-4         & 133.10+83.70  & 1352.07 & 881.00 & 128.18+113.70 & 1924.35 & 131.30+150.27 & 64.55  & 129.84+176.65 & 855.30   \\
PC-3           & 91.26+70.05   & 1249.30 & 879.76 & 90.55+64.84   & 1484.25 & 90.46+110.26  & 51.52  & 90.70+98.88   & 613.19   \\
SW-620         & 133.80+102.96 & 1248.39 & 875.52 & 136.75+99.17  & 2076.36 & 133.42+158.47 & 75.68  & 133.50+153.30 & 873.41   \\
NCI-H23        & 130.20+108.73 & 1301.09 & 883.09 & 130.63+93.86  & 2091.18 & 131.54+134.20 & 63.68  & 131.07+143.53 & 896.41   \\
OVCAR-8        & 131.48+98.46  & 1351.17 & 890.68 & 132.62+102.36 & 2077.41 & 132.07+168.57 & 61.41  & 131.23+130.47 & 986.01   \\
P388           & 120.73+103.09 & 1258.91 & 884.15 & 120.43+90.74  & 2140.75 & 122.53+145.56 & 60.47  & 121.12+135.97 & 965.61   \\
SF-295         & 130.92+84.81  & 1359.48 & 879.52 & 131.26+110.89 & 2063.71 & 130.77+168.45 & 66.25  & 130.43+150.38 & 915.04   \\
SN12C          & 129.79+102.16 & 1334.82 & 914.87 & 129.62+87.23  & 2044.62 & 130.21+148.75 & 68.58  & 128.95+134.45 & 933.55   \\
UACC257        & 128.63+84.19  & 1387.70 & 930.43 & 128.39+88.05  & 2053.46 & 129.04+138.45 & 58.83  & 129.65+142.53 & 891.95   \\
PROTEINS\_full & 9.03+6.67     & 206.50  & 202.65 & 8.50+8.86     & 69.30   & 8.62+7.98     & 4.58   & 8.84+7.63     & 30.25    \\
DBLP\_v1       & 38.86+62.06   & 1335.42 & 897.90 & 36.99+61.50   & 1123.57 & 39.22+99.74   & 30.51  & 39.15+105.83  & 155.67  \\ \hline \hline
\end{tabular}}
\end{table}
For the Preprocess function, we first analyze the time complexity of the matrix multiplication. Since we utilize sparse matrices to conduct the experiment, the time complexity of the multiplication is $O(\text{nnz}(G.\vect{D}^{-\frac{1}{2}})*\text{nnz}(G.\vect{A})+\text{nnz}(G.\vect{D}^{-\frac{1}{2}}*G.\vect{A})*\text{nnz}(G.\vect{D}^{-\frac{1}{2}})$, where nnz means non-zero entries of the matrix. Then, by adopting EVD to only keep the top-$k_l$
largest and top-$k_s$ smallest eigenvalues, the time complexity can be $O(n*(k_l^2+k_s^2) + m*(k_l+k_s))$, where $n,m$ is the number of nodes/edges. As shown in Algorithm \ref{alg:fracaug}, Preprocess can be called before the training process, and thus it won't burden the training or inference of our FracAug. 

Then, we analyze the time complexity of $\text{FGG}$ for each graph. As presented in Algorithm \ref{alg:fgg}, we perform sparse matrix multiplication for every sample $H_l$ and $H_s$ times. Besides, since $G.\vect{\Lambda}$ is a diagonal matrix, the time complexity of multiplying $G.\vect{\Lambda}$ is the same as that of multiplying $G.\vect{\Lambda}^\alpha$. Therefore, the total time complexity of $\text{FGG}$ is $O(\text{FGG})=O(\text{nnz}(G.\vect{U}_l)*k_l+\text{nnz}(G.\vect{U}_l*G.\vect{\Lambda}_l)*\text{nnz}(G.\vect{U}_l^T)+\text{nnz}(G.\vect{U}_s)*k_s+\text{nnz}(G.\vect{U}_s*G.\vect{\Lambda}_s)*\text{nnz}(G.\vect{U}_s^T))$. 

Next, we analyze the time complexity of $\text{WDML}$ in Algorithm \ref{alg:wdml}. Assuming that we only have one sample in $\mathcal{D}_{train}$, then the time complexity of WDML is $O(\text{WDML})=O(d)$, where $d$ is the dimension of the generated graph embedding $\vect{o}$. 

Moreover, as shown in Algorithm \ref{alg:mvp}, in MVP, we only need to see if the probability predicted by the given GNN satisfies the criterion, so the time complexity for MVP is $O(\text{MVP})=O(1)$.

Finally, in Algorithm \ref{alg:fracaug}, we combine all the time complexities together within one training epoch of the given GNN $f$, assuming the time complexity of $f$ for each sample is $O(f)$, then we have the total complexity as $O(e_{\text{FGG}}*(O(\text{FGG})+O(\text{WDML})+O(f))*N_{train}+(O(\text{FGG})+O(\text{MVP})+O(f))*(N_{val}+N_{test})$, where $N_{val}, N_{test}$ represent the number of samples in $\mathcal{D}_{val}$ and $\mathcal{D}_{test}$, respectively. 

In practice, we set $e_{FGG}$ to 10 and $e_{aug}$ to 25, which can reduce the computational cost, and FGG can still converge. According to the final complexity, we can see the dominant factor within each epoch is $O(e_{FGG} * O(f) * N_{train}+O(f)*(N_{val}+N_{test}))$. For such a factor, we need to calculate it in total $\frac{e_f-e_{warmup}}{e_{aug}}*e_{FGG}$ times, which is much less than the original training epoch of $f$. Hence, the increase in time complexity will not be the limitation of our FracAug in real applications. 

In Table \ref{tab:runningtime}, we present a detailed runtime comparison between FracAug and several leading graph augmentation methods. For each technique, we decompose the total computational cost into a one-time preprocessing phase, performed once per dataset, and the subsequent training time measured over multiple epochs. While some baselines require repeated feature perturbations or costly online sampling at every iteration, FracAug’s eigenvalue decomposition is only performed during preprocessing. As a result, the per-epoch training overhead of FracAug remains on par with, or even below, that of competing approaches, despite leveraging additional spectral information to boost anomaly detection performance.

Crucially, these efficiency gains do not come at the expense of detection performance. Across all datasets and baseline comparisons, FracAug consistently delivers state-of-the-art AUROC, AUPRC, and F1-score results while maintaining competitive total runtimes. By amortizing the heavier spectral computations over the entire training cycle and by implementing optimized matrix operations, FracAug strikes an effective balance between computational tractability and augmentation quality. This combination of speed and performance underscores the practical value of our method: practitioners can readily adopt FracAug for real graph applications without incurring prohibitive time costs.

\section{Experimental Settings}
\label{app:experimentsetting}

\begin{table}[t]
\caption{Hyperparameters of 12 datasets based on GIN.}
\small
\centering
\scalebox{0.88}{
\setlength\tabcolsep{1pt}
\label{tab:hyperparametersetting}
\begin{tabular}{ccccccccccccc} \hline \hline
Datasets         & MCF-7 & MOLT-4 & PC-3 & SW-620 & NCI-H23 & OVCAR-8 & P388 & SF-295 & SN12C & UACC257 & PROTEINS\_full & DBLP\_v1 \\ \hline
$k_l$            & 4     & 4      & 4    & 4      & 3       & 3       & 4    & 3      & 4     & 4       & 3              & 4        \\
$H_l$            & 4     & 3      & 3    & 3      & 3       & 3       & 4    & 4      & 3     & 4       & 4              & 4        \\
$k_s$            & 3     & 4      & 3    & 3      & 4       & 3       & 4    & 3      & 3     & 4       & 3              & 3        \\
$H_s$            & 4     & 4      & 3    & 3      & 3       & 3       & 4    & 3      & 3     & 4       & 4              & 3        \\
$e_{warmup}$ & 50    & 25     & 50   & 50     & 25      & 25      & 50   & 50     & 25    & 50      & 50             & 25      \\ \hline \hline
\end{tabular}}
\end{table}

Table \ref{tab:hyperparametersetting} provides a comprehensive list of our hyperparameters based on GIN. We use grid search to train FracAug, which yields the best sum of AUROC, AUPRC, and F1-score on the validation set, and report the corresponding test performance. Specifically, $k_l, H_l, k_s, H_s$ range from the set $\{3, 4\}$, and the number of warmup epoch for GNN is selected from $\{25, 50\}$, to reduce the cost of search. In the next Section \ref{app:hyperparameteranalysis}, we further analyze the influence of $k_l, H_l, k_s, H_s$ on the AUROC, AUPRC, and F1-score of different datasets based on GIN. As for the experimental environment, we conduct all the experiments on an NVIDIA Quadro RTX 8000 for a fair comparison.

\section{Hyperarameter Analysis}
\label{app:hyperparameteranalysis}

\begin{table}[t]
\caption{Varying $k_l$-$k_s$-$H_l$-$H_s$ on different datasets based on GIN.}
\small
\centering
\scalebox{1.1}{
\setlength\tabcolsep{1.5pt}
\label{tab:parameterkt}
\begin{tabular}{c|ccc|ccc} \hline \hline
Datasets & \multicolumn{3}{c|}{PROTEINS\_full} & \multicolumn{3}{c}{DBLP\_v1} \\ 
$k_l$-$k_s$-$H_l$-$H_s$  & AUROC     & AUPRC     & F1-score   & AUROC   & AUPRC   & F1-score \\  \hline
3-3-3-3  & 0.6174    & 0.6298    & 0.6187     & 0.7950  & 0.8514  & 0.7947   \\
3-3-3-4  & 0.6141    & 0.6327    & 0.6142     & 0.7972  & 0.8572  & 0.7955   \\
3-3-4-3  & 0.6103    & 0.6291    & 0.6104     & 0.7995  & 0.8546  & 0.7992   \\
3-3-4-4  & 0.6174    & 0.6358    & 0.6175     & 0.7925  & 0.8486  & 0.7925   \\
3-4-3-3  & 0.6174    & 0.6298    & 0.6187     & 0.7950  & 0.8514  & 0.7947   \\
3-4-3-4  & 0.6082    & 0.6304    & 0.6072     & 0.7972  & 0.8572  & 0.7955   \\
3-4-4-3  & 0.6103    & 0.6291    & 0.6104     & 0.7995  & 0.8546  & 0.7992   \\
3-4-4-4  & 0.6094    & 0.6210    & 0.6104     & 0.7982  & 0.8524  & 0.7982   \\
4-3-3-3  & 0.6174    & 0.6298    & 0.6187     & 0.7885  & 0.8509  & 0.7867   \\
4-3-3-4  & 0.6124    & 0.6284    & 0.6133     & 0.7972  & 0.8538  & 0.7967   \\
4-3-4-3  & 0.6161    & 0.6295    & 0.6174     & 0.8044  & 0.8626  & 0.8028   \\
4-3-4-4  & 0.6138    & 0.6316    & 0.6142     & 0.8007  & 0.8570  & 0.8000   \\
4-4-3-3  & 0.6174    & 0.6298    & 0.6187     & 0.7972  & 0.8579  & 0.7953   \\
4-4-3-4  & 0.6161    & 0.6295    & 0.6174     & 0.7994  & 0.8560  & 0.7987   \\
4-4-4-3  & 0.6108    & 0.6334    & 0.6095     & 0.8015  & 0.8617  & 0.7996   \\
4-4-4-4  & 0.6094    & 0.6210    & 0.6104     & 0.8008  & 0.8577  & 0.7999  \\ \hline \hline
\end{tabular}}
\end{table}
\begin{figure}[t]
\centering
  \begin{small}

    \begin{tabular}{ccc}
        \hspace{-6mm}
        \includegraphics[height=48mm]{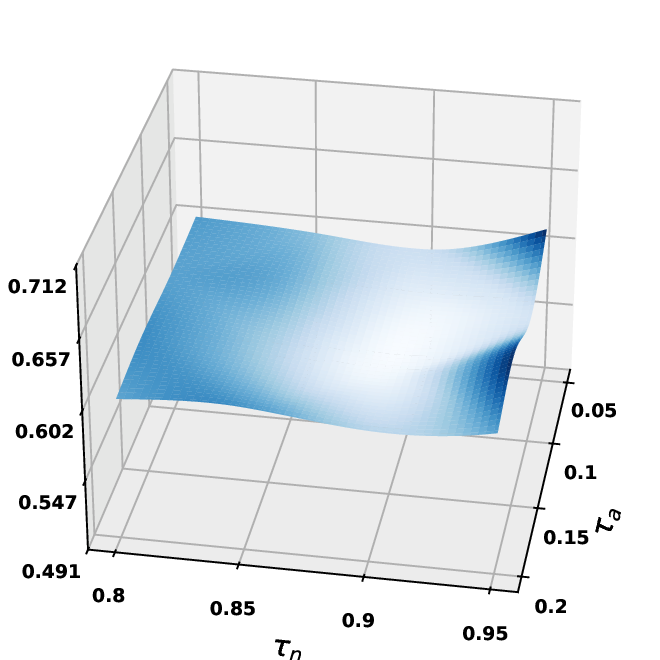} &
        \hspace{-6mm}
        \includegraphics[height=48mm]{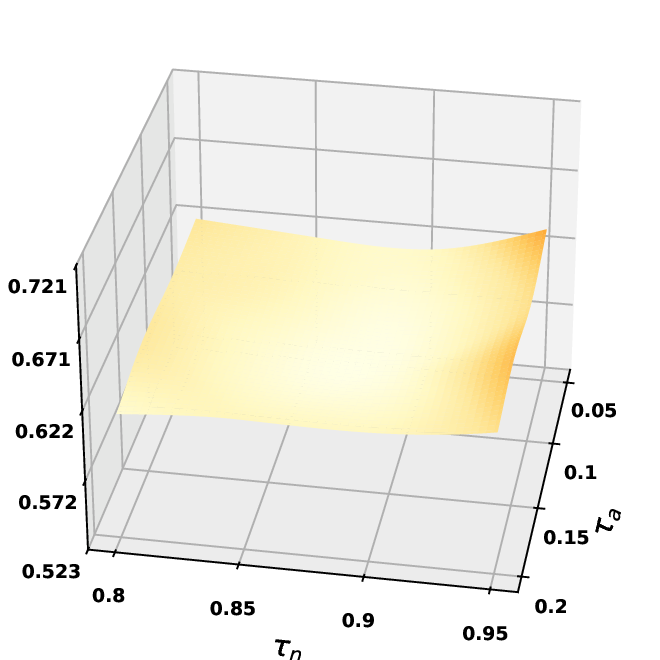} &
        \hspace{-6mm}
        \includegraphics[height=48mm]{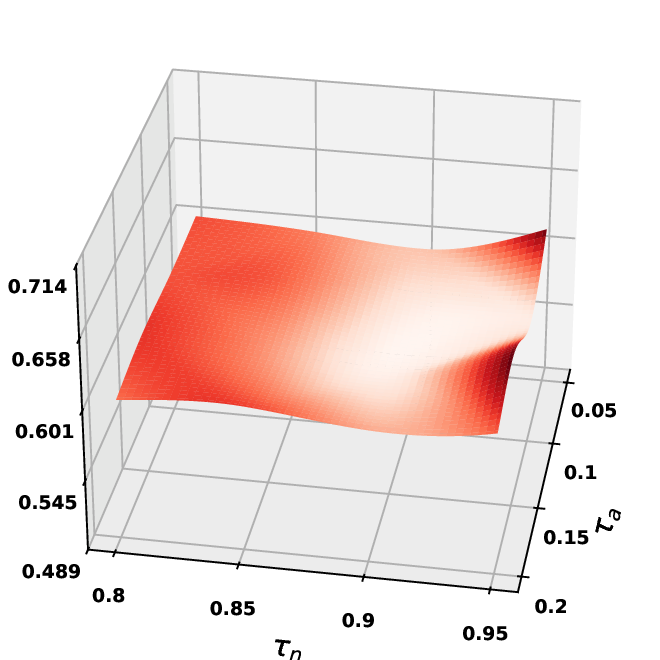} \\ [-0mm]
        \hspace{-9mm}
        (a) AUROC & 
        \hspace{-9mm}
        (b) AUPRC &
        \hspace{-9mm}
        (c) F1-score \\ 
    \end{tabular}
    \vspace{-2mm}
    \caption{Varying $\tau_a$ and $\tau_n$ for PROTEINS\_full based on GIN.}
    \label{fig:hyperparameterprotiens}
  \vspace{-8mm}
  \end{small}
\end{figure}
\begin{figure}[t]
\centering
  \begin{small}

    \begin{tabular}{ccc}
        \hspace{-6mm}
        \includegraphics[height=48mm]{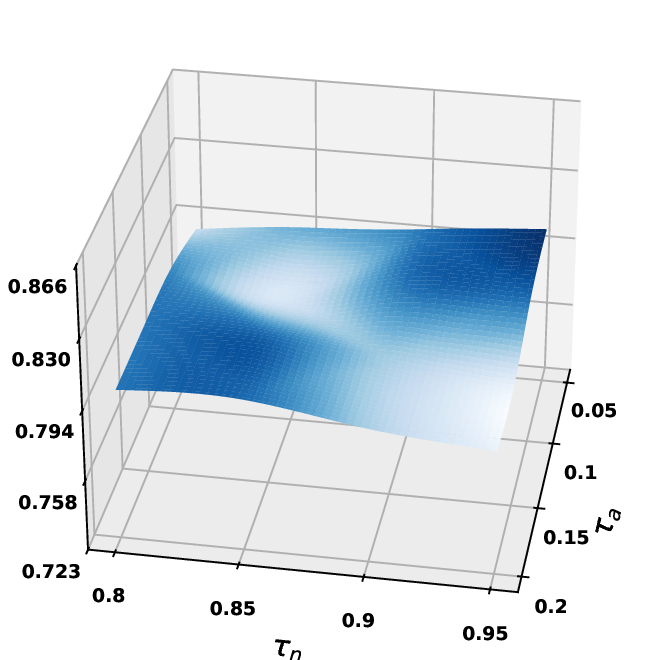} &
        \hspace{-6mm}
        \includegraphics[height=48mm]{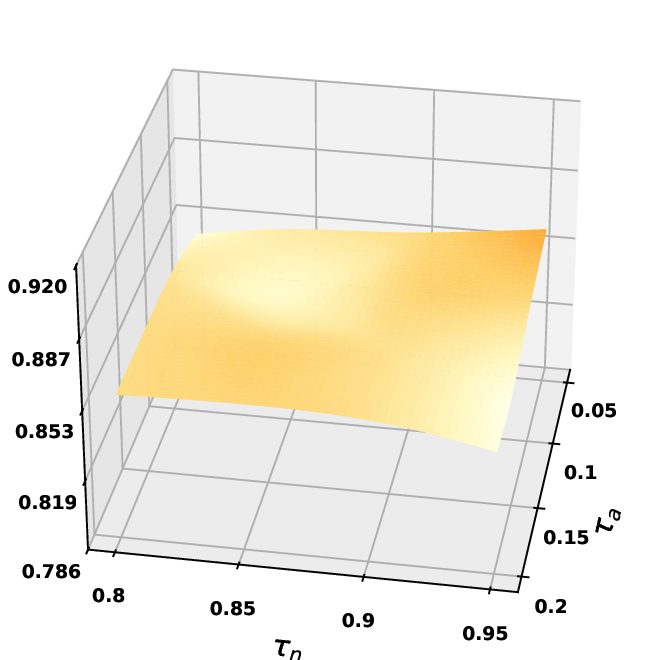} &
        \hspace{-6mm}
        \includegraphics[height=48mm]{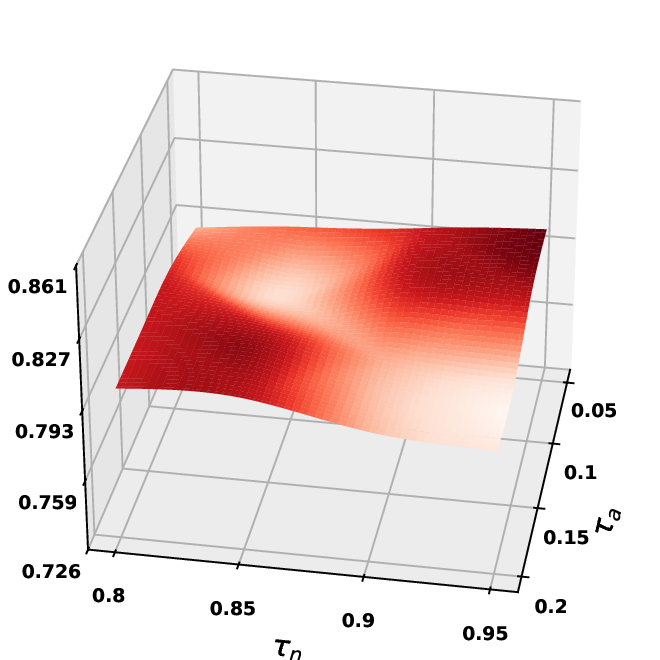} \\ [-0mm]
        \hspace{-9mm}
        (a) AUROC & 
        \hspace{-9mm}
        (b) AUPRC &
        \hspace{-9mm}
        (c) F1-score \\ 
    \end{tabular}
    \vspace{-2mm}
    \caption{Varying $\tau_a$ and $\tau_n$ for DBLP\_v1 based on GIN.}
    \label{fig:hyperparameterdblp}
  \vspace{-2mm}
  \end{small}
\end{figure}

Table \ref{tab:parameterkt} presents a systematic exploration of FracAug’s performance, measured in AUROC, AUPRC, and F1-score, when varying $k_l, k_s, H_l, H_s$ between 3 and 4. Specifically, $k_l, k_s$ denote the top-$k_l$ largest and top-$k_s$ eigenvalues generated by EVD, while $H_l, H_s$ denote the number of learnable fractional powers of the matrix for the largest and smallest eigenvalues. By sweeping each of these four parameters, we generate a compact grid of 16 configurations. As detailed in Table \ref{tab:hyperparametersetting}, each evaluation metric prefers a slightly different quadruple of ($k_l, k_s, H_l, H_s$), but more importantly, Table \ref{tab:parameterkt} reveals that the detection performance barely wavers across all measured combinations. This robustness not only validates the spectral augmentation strategy at the heart of FracAug but also suggests that practitioners can avoid laborious hyperparameter sweeps without sacrificing anomaly detection performance.

In a parallel study, Figures \ref{fig:hyperparameterprotiens} and \ref{fig:hyperparameterdblp} examine the sensitivity of FracAug to the pseudo-labeling thresholds $\tau_n$ and $\tau_a$. Here, $\tau_n$ specifies the percentile above which a node is considered “normal,” and $\tau_a$ the percentile below which it is flagged as “anomalous.” By varying $\tau_n$ from $0.8$ to $0.95$ and $\tau_a$ from $0.05$ to $0.2$, we again explore 16 threshold pairs on each dataset, logging the resulting AUROC, AUPRC, and F1-score for every pair. Remarkably, all three metrics remain essentially flat throughout this entire range, indicating that FracAug’s pseudo-labeling module is forgiving of moderate threshold choices. Leveraging this newfound stability, we adopt the pair $\tau_n=0.05$ and $\tau_a=0.95$ as our default across all datasets, thereby slashing the computational overhead of threshold tuning without meaningfully affecting detection performance.

\section{Ablation Study}
\label{app:ablationstudy}

\begin{table}[t]
\caption{Ablation study.}
\small
\centering
\scalebox{1}{
\setlength\tabcolsep{1.5pt}
\label{tab:ablationstudy}
\begin{tabular}{cc|c
>{\columncolor[HTML]{DCDCDC}}c cccc} \hline \hline
Datasets       & Metrics  & GIN     & \multicolumn{1}{l}{\cellcolor[HTML]{ffffff}{\scriptsize +FA}} & w/o largest & w/o smallest & w/o WDML & w/o MVP \\ \hline
\multirow{3}{*}{MCF-7}          & AUROC    & 0.5867 & 0.5976               & 0.5848      & 0.5889       & 0.5860   & 0.5835  \\
                                & AUPRC    & 0.2830 & 0.2971               & 0.2842      & 0.2882       & 0.2813   & 0.2790  \\
                                & F1-score & 0.5366 & 0.5421               & 0.5317      & 0.5351       & 0.5372   & 0.5350  \\ \hline
\multirow{3}{*}{MOLT-4}         & AUROC    & 0.5733 & 0.5854               & 0.5770      & 0.5760       & 0.5772   & 0.5754  \\
                                & AUPRC    & 0.2830 & 0.3001               & 0.2974      & 0.2862       & 0.2981   & 0.2881  \\
                                & F1-score & 0.5072 & 0.5103               & 0.5001      & 0.5085       & 0.4998   & 0.5060  \\ \hline
\multirow{3}{*}{PC-3}           & AUROC    & 0.5969 & 0.6119               & 0.5963      & 0.6018       & 0.6026   & 0.5975  \\
                                & AUPRC    & 0.2797 & 0.2893               & 0.2797      & 0.2815       & 0.2806   & 0.2780  \\
                                & F1-score & 0.5063 & 0.5205               & 0.5055      & 0.5124       & 0.5145   & 0.5092  \\ \hline
\multirow{3}{*}{SW-620}         & AUROC    & 0.5938 & 0.6004               & 0.5941      & 0.5938       & 0.5949   & 0.5930  \\
                                & AUPRC    & 0.2776 & 0.2813               & 0.2737      & 0.2778       & 0.2800   & 0.2697  \\
                                & F1-score & 0.5090 & 0.5155               & 0.5132      & 0.5089       & 0.5082   & 0.5155  \\ \hline
\multirow{3}{*}{NCI-H23}        & AUROC    & 0.5897 & 0.5968               & 0.5893      & 0.5911       & 0.5866   & 0.5925  \\
                                & AUPRC    & 0.2566 & 0.2659               & 0.2564      & 0.2633       & 0.2614   & 0.2656  \\
                                & F1-score & 0.5059 & 0.5073               & 0.5054      & 0.5013       & 0.4968   & 0.5013  \\ \hline
\multirow{3}{*}{OVCAR-8}        & AUROC    & 0.5935 & 0.5963               & 0.5911      & 0.5918       & 0.5911   & 0.5904  \\
                                & AUPRC    & 0.2573 & 0.2612               & 0.2579      & 0.2565       & 0.2573   & 0.2605  \\
                                & F1-score & 0.5118 & 0.5123               & 0.5074      & 0.5100       & 0.5081   & 0.5038  \\ \hline
\multirow{3}{*}{P388}           & AUROC    & 0.5565 & 0.5913               & 0.5620      & 0.5708       & 0.5730   & 0.5599  \\
                                & AUPRC    & 0.2850 & 0.3309               & 0.2917      & 0.3050       & 0.3074   & 0.3014  \\
                                & F1-score & 0.4468 & 0.4491               & 0.4482      & 0.4470       & 0.4470   & 0.4371  \\ \hline
\multirow{3}{*}{SF-295}         & AUROC    & 0.5844 & 0.6076               & 0.5971      & 0.5949       & 0.5920   & 0.5996  \\
                                & AUPRC    & 0.2766 & 0.2832               & 0.2723      & 0.2716       & 0.2653   & 0.2790  \\
                                & F1-score & 0.4803 & 0.5047               & 0.5001      & 0.4975       & 0.4994   & 0.4973  \\ \hline
\multirow{3}{*}{SN12C}          & AUROC    & 0.5995 & 0.6079               & 0.5993      & 0.5963       & 0.5910   & 0.5990  \\
                                & AUPRC    & 0.2696 & 0.2746               & 0.2685      & 0.2666       & 0.2632   & 0.2675  \\
                                & F1-score & 0.5030 & 0.5110               & 0.5041      & 0.5013       & 0.4971   & 0.5046  \\ \hline
\multirow{3}{*}{UACC257}        & AUROC    & 0.5877 & 0.6015               & 0.5939      & 0.5854       & 0.5938   & 0.5914  \\
                                & AUPRC    & 0.2480 & 0.2598               & 0.2517      & 0.2527       & 0.2585   & 0.2586  \\
                                & F1-score & 0.4906 & 0.4983               & 0.4956      & 0.4835       & 0.4890   & 0.4858  \\ \hline
\multirow{3}{*}{PROTEINS\_full} & AUROC    & 0.5799 & 0.6174               & 0.5986      & 0.5842       & 0.5990   & 0.5854  \\
                                & AUPRC    & 0.6259 & 0.6358               & 0.6111      & 0.5988       & 0.6205   & 0.6042  \\
                                & F1-score & 0.5679 & 0.6175               & 0.5995      & 0.5848       & 0.5976   & 0.5857  \\ \hline
\multirow{3}{*}{DBLP\_v1}       & AUROC    & 0.6231 & 0.8044               & 0.7615      & 0.7828       & 0.7762   & 0.7628  \\
                                & AUPRC    & 0.7201 & 0.8626               & 0.8301      & 0.8411       & 0.8374   & 0.8294  \\
                                & F1-score & 0.5996 & 0.8028               & 0.7594      & 0.7829       & 0.7760   & 0.7619 \\  \hline \hline
\end{tabular}}
\end{table}

To examine the effectiveness of each component in FracAug, we conduct ablation study on 12 datasets based on GIN, which is shown in Table \ref{tab:ablationstudy}. Specifically, the gray column represents the performance of GIN with FracAug, w/o largest and w/o smallest denotes removing the fractional graphs generated by top-$k_l$ largest and top-$k_s$ smallest eigenvalues, respectively, w/o WDML means replacing our proposed WDML with weighted cross-entropy loss, and w/o MVP pseudo-labels samples in $\mathcal{D}_{val} \cup \mathcal{D}_{test}$ using only the predicted probability of original graphs. As shown in Table \ref{tab:ablationstudy}, FracAug consistently outperforms its variants, w/o largest, w/o smallest, w/o WDML, and w/o MVP, which demonstrates the benefits of these components. 

\section{Margin Loss Comparison}
\label{app:lossfunctioncomoparson}

\begin{table}[t]
\caption{Comparison of different margin loss. }
\small
\centering
\scalebox{1}{
\setlength\tabcolsep{1.5pt}
\label{tab:marginlosscomparison}
\begin{tabular}{cc|c
>{\columncolor[HTML]{DCDCDC}}c ccc} \hline \hline
Datasets       & Metrics  & GIN     & \multicolumn{1}{l}{\cellcolor[HTML]{ffffff}{\scriptsize +FA}} & Softmax & LMCL   & LDAM   \\ \hline 
\multirow{3}{*}{MCF-7}          & AUC     & 0.5867 & 0.5976 & 0.5844  & 0.5880 & 0.5844 \\
                                & AUPRC   & 0.2830 & 0.2971 & 0.2781  & 0.2847 & 0.2796 \\
                                & MF1     & 0.5366 & 0.5421 & 0.5378  & 0.5372 & 0.5360 \\ \hline 
\multirow{3}{*}{MOLT-4}         & AUC     & 0.5733 & 0.5854 & 0.5760  & 0.5797 & 0.5751 \\
                                & AUPRC   & 0.2830 & 0.3001 & 0.2969  & 0.2952 & 0.2857 \\
                                & MF1     & 0.5072 & 0.5103 & 0.4991  & 0.5059 & 0.5076 \\ \hline 
\multirow{3}{*}{PC-3}           & AUC     & 0.5969 & 0.6119 & 0.6021  & 0.6037 & 0.6001 \\
                                & AUPRC   & 0.2797 & 0.2893 & 0.2809  & 0.2778 & 0.2769 \\
                                & MF1     & 0.5063 & 0.5205 & 0.5134  & 0.5195 & 0.5144 \\ \hline 
\multirow{3}{*}{SW-620}         & AUC     & 0.5938 & 0.6004 & 0.5947  & 0.5936 & 0.5931 \\
                                & AUPRC   & 0.2776 & 0.2813 & 0.2779  & 0.2768 & 0.2720 \\
                                & MF1     & 0.5090 & 0.5155 & 0.5100  & 0.5092 & 0.5133 \\ \hline 
\multirow{3}{*}{NCI-H23}        & AUC     & 0.5897 & 0.5968 & 0.5913  & 0.5896 & 0.5887 \\
                                & AUPRC   & 0.2566 & 0.2659 & 0.2650  & 0.2612 & 0.2622 \\
                                & MF1     & 0.5059 & 0.5073 & 0.5001  & 0.5012 & 0.4990 \\ \hline 
\multirow{3}{*}{OVCAR-8}        & AUC     & 0.5935 & 0.5963 & 0.5905  & 0.5890 & 0.5932 \\
                                & AUPRC   & 0.2573 & 0.2612 & 0.2557  & 0.2570 & 0.2579 \\
                                & MF1     & 0.5118 & 0.5123 & 0.5087  & 0.5051 & 0.5107 \\ \hline 
\multirow{3}{*}{P388}           & AUC     & 0.5565 & 0.5913 & 0.5864  & 0.5653 & 0.5602 \\
                                & AUPRC   & 0.2850 & 0.3309 & 0.3265  & 0.3080 & 0.2901 \\
                                & MF1     & 0.4468 & 0.4491 & 0.4465  & 0.4377 & 0.4469 \\ \hline 
\multirow{3}{*}{SF-295}         & AUC     & 0.5844 & 0.6076 & 0.5943  & 0.5898 & 0.5955 \\
                                & AUPRC   & 0.2766 & 0.2832 & 0.2664  & 0.2712 & 0.2715 \\
                                & MF1     & 0.4803 & 0.5047 & 0.5017  & 0.4909 & 0.4985 \\ \hline 
\multirow{3}{*}{SN12C}          & AUC     & 0.5995 & 0.6079 & 0.5914  & 0.5955 & 0.6004 \\
                                & AUPRC   & 0.2696 & 0.2746 & 0.2661  & 0.2601 & 0.2707 \\
                                & MF1     & 0.5030 & 0.5110 & 0.4950  & 0.5068 & 0.5034 \\ \hline 
\multirow{3}{*}{UACC257}        & AUC     & 0.5877 & 0.6015 & 0.5905  & 0.5899 & 0.5935 \\
                                & AUPRC   & 0.2480 & 0.2598 & 0.2584  & 0.2581 & 0.2557 \\
                                & MF1     & 0.4906 & 0.4983 & 0.4849  & 0.4844 & 0.4912 \\ \hline 
\multirow{3}{*}{PROTEINS\_full} & AUC     & 0.5799 & 0.6174 & 0.6002  & 0.5937 & 0.6051 \\
                                & AUPRC   & 0.6259 & 0.6358 & 0.6235  & 0.6078 & 0.6231 \\
                                & MF1     & 0.5679 & 0.6175 & 0.5987  & 0.5945 & 0.6051 \\ \hline 
\multirow{3}{*}{DBLP\_v1}       & AUC     & 0.6231 & 0.8044 & 0.7776  & 0.7834 & 0.7874 \\
                                & AUPRC   & 0.7201 & 0.8626 & 0.8383  & 0.8463 & 0.8502 \\
                                & MF1     & 0.5996 & 0.8028 & 0.7774  & 0.7817 & 0.7855 \\  \hline  \hline 
\end{tabular}}
\end{table}

Next, we investigate the performance of FracAug with different margin loss as stated in Section \ref{subsec:weightedistanceawaremarginloss}, by conducting comparison on 12 datasets based on GIN. As shown in Table \ref{tab:marginlosscomparison}, FracAug consistently outperforms its variants, Softmax, LMCL, and LDAM. Such a phenomenon demonstrates that sample-specific decision boundaries can be more effective than fixed margins for GAD tasks, aligning with our analysis in Section \ref{subsec:weightedistanceawaremarginloss}. 

\section{Performance with More Training Data}
\label{app:performancewithmoredata}

\begin{figure}[t]
\centering

\includegraphics[height=38mm]{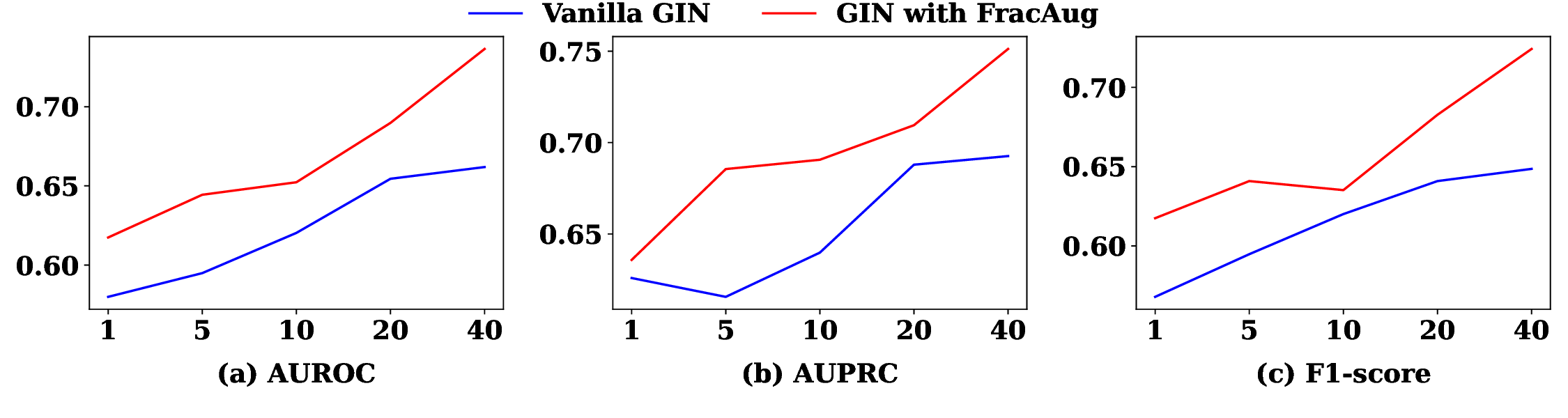}
\vspace{-8mm}
\caption{Varying training size (\%) for PROTEINS\_full.}
\label{fig:trainszproteins}
\end{figure}
\begin{figure}[t]
\centering

\includegraphics[height=38mm]{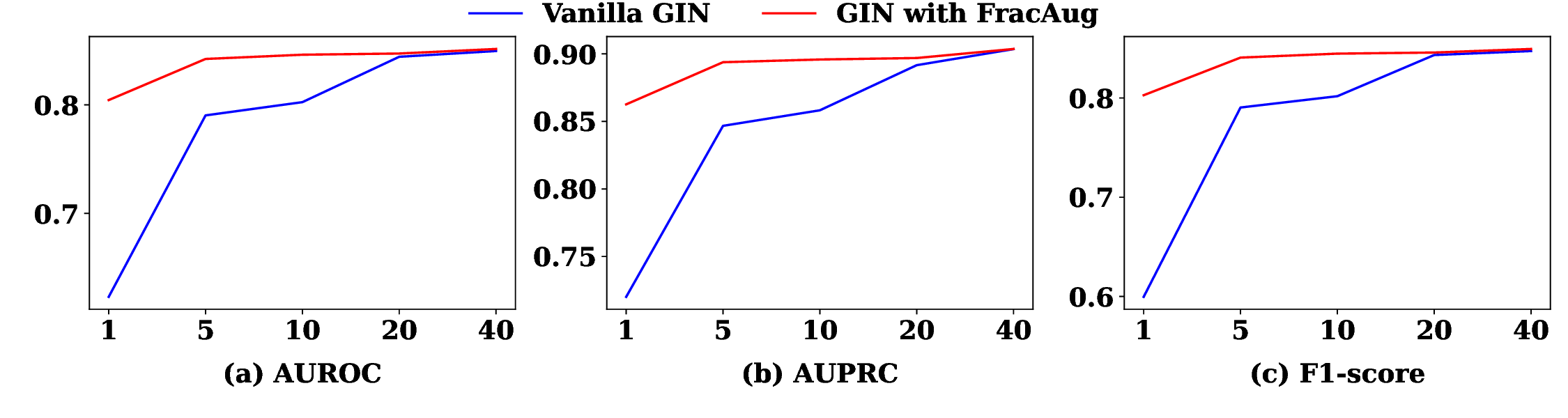}
\vspace{-8mm}
\caption{Varying training size (\%) for DBLP\_v1.}
\label{fig:trainszdblp}
\end{figure}

To further demonstrate the superior ability of our proposed FracAug, we conduct experiments on PROTEINS\_full and DBLP\_v1, varying by the training size (\%), as shown in Figures \ref{fig:trainszproteins} and \ref{fig:trainszdblp}. As we can see, the red lines, which represent the performance of GIN with FracAug, are always on top of the figures, which demonstrates that with more training data, our FracAug can still boost the performance of the given baseline consistently in terms of AUROC, AUPRC, and F1-score. To sum up, such experiments further prove that our FracAug is a more general and practical framework.

\section{Learned Parameters}
\label{app:learnedparameter}

\begin{table}[t]
\caption{Learned parameters}
\small
\centering
\scalebox{0.87}{
\setlength\tabcolsep{1pt}
\label{tab:learnedparameters}
\begin{tabular}{c|cccccccccccc} \hline \hline
Datasets                           & MCF-7  & MOLT-4 & PC-3   & SW-620 & NCI-H23 & OVCAR-8 & P388   & SF-295 & SN12C  & UACC257 & PROTEINS\_full & DBLP\_v1 \\ \hline
\multirow{4}{*}{$\vect{\alpha}_l$}      & 0.9062 & 1.4187 & 1.0862 & 2.186  & 0.4991  & 0.8986  & 2.8806 & 0.8845 & 2.7404 & 1.6178  & 1.1407         & 1.8663   \\
                                   & 2.0063 & 2.2813 & 1.9282 & 1.5108 & 0.3535  & 0.1262  & 2.9620 & 2.9038 & 0.5877 & 1.9339  & 0.6885         & 1.1374   \\
                                   & 1.4770 & 1.7911 & 1.9928 & 1.8073 & 2.3672  & 1.1480  & 2.3129 & 2.3939 & 1.0219 & 2.4874  & 2.5432         & 1.1843   \\
                                   & 1.4269 & -      & -      & -      & -       & -       & 1.2674 & 2.4786 & -      & 2.2634  & 1.4364         & 2.8254   \\ \hline
\multirow{4}{*}{$\vect{\omega}_l$} & 0.2921 & 0.2655 & 0.3450 & 0.2667 & 0.2793  & 0.3476  & 0.1639 & 0.3051 & 0.2385 & 0.1165  & 0.3842         & 0.1685   \\
                                   & 0.3108 & 0.3101 & 0.3349 & 0.5068 & 0.4612  & 0.3615  & 0.3654 & 0.1747 & 0.3833 & 0.2806  & 0.1535         & 0.1714   \\
                                   & 0.1549 & 0.4244 & 0.3202 & 0.2266 & 0.2595  & 0.2909  & 0.2728 & 0.2145 & 0.3782 & 0.3028  & 0.2927         & 0.4004   \\
                                   & 0.2422 & -      & -      & -      & -       & -       & 0.1979 & 0.3057 & -      & 0.3000  & 0.1696         & 0.2597   \\ \hline
\multirow{4}{*}{$\vect{\alpha}_s$}      & 2.3128 & 1.9621 & 2.8048 & 1.5579 & 2.4828  & 0.5030  & 1.5344 & 1.6912 & 2.9331 & 0.8959  & 1.1290         & 1.7201   \\
                                   & 2.1539 & 0.4087 & 1.1722 & 1.8088 & 2.1366  & 1.4235  & 1.552  & 1.9896 & 1.5150 & 1.9589  & 0.1467         & 2.8101   \\
                                   & 0.7485 & 2.9168 & 1.6098 & 0.4474 & 2.2668  & 2.6830  & 2.0318 & 1.8741 & 2.9352 & 2.8844  & 0.5492         & 2.0935   \\
                                   & 2.5362 & 0.2449 & -      & -      & -       & -       & 2.7135 & -      & -      & 0.1148  & 0.0124         & -        \\ \hline
\multirow{4}{*}{$\vect{\omega}_s$} & 0.3042 & 0.2821 & 0.3910 & 0.3756 & 0.3292  & 0.3384  & 0.1696 & 0.4052 & 0.4554 & 0.2916  & 0.2407         & 0.2517   \\
                                   & 0.2569 & 0.2112 & 0.2750 & 0.4245 & 0.3630  & 0.3436  & 0.4001 & 0.2386 & 0.2343 & 0.2539  & 0.3285         & 0.4758   \\
                                   & 0.2764 & 0.1519 & 0.3340 & 0.1999 & 0.3078  & 0.3180  & 0.2027 & 0.3562 & 0.3103 & 0.3122  & 0.2524         & 0.2725   \\
                                   & 0.1625 & 0.3548 & -      & -      & -       & -       & 0.2277 & -      & -      & 0.1423  & 0.1785         & -        \\ \hline
\multirow{2}{*}{$\vect{\omega}$}   & 0.4634 & 0.6795 & 0.6481 & 0.5579 & 0.5263  & 0.4541  & 0.4408 & 0.5589 & 0.5724 & 0.5309  & 0.7134         & 0.5174   \\
                                   & 0.5366 & 0.3205 & 0.3519 & 0.4421 & 0.4737  & 0.5459  & 0.5592 & 0.4411 & 0.4276 & 0.4691  & 0.2866         & 0.4826  \\ \hline \hline
\end{tabular}}
\end{table}

Here, we present the learned parameters based on GIN, $\vect{\alpha}_l$, $\vect{\omega}_l$, $\vect{\alpha}_s$, $\vect{\omega}_s$, and $\vect{\omega}$ in Table \ref{tab:learnedparameters}. Specifically, $\vect{\alpha}_l, \vect{\alpha}_s$ represent the learned fractional powers for the largest and smallest eigenvalues, respectively, $\vect{\omega}_l, \vect{\omega}_s$ denote the corresponding learned coefficients for the fractional graphs, and $\vect{\omega}$ means the balanced coefficient between the group of fractional graphs from the largest and smallest eigenvalues. Recap from Section \ref{subsec:fractionalgraphgenerator}, we utilize $H_l$ and $H_s$ to control the number of combined fractional graphs from different eigenvalues, so the number of entries in $\vect{\alpha}_l, \vect{\omega}_l$ is equal to $H_l$ and that of $\vect{\alpha}_s, \vect{\omega}_s$ is the same as $H_s$. Since we only consider the largest and smallest eigenvalue groups, the size of $\vect{\omega}$ should be 2. Note that, according to Table \ref{tab:hyperparametersetting}, the optimal $H_l, H_s$ varies by datasets, so the number of entries in Table \ref{tab:learnedparameters} will also be different, where "-" represents no such entry in the vector. As shown in Table \ref{tab:learnedparameters}, the learned powers vary within the range of $(0, 3)$, which aligns with the parameter settings of common GNNs, as deeper GNNs may result in over-smoothing and shallower GNNs will lead to under-fitting. This experiment further shows the rationality of our proposed FracAug.

\end{document}